\documentclass[11pt,a4paper]{article}
\usepackage[hyperref]{acl2021}
\usepackage{times}
\usepackage{latexsym}
\usepackage[textsize=tiny]{todonotes}

\usepackage[utf8]{inputenc}
\usepackage[T1]{fontenc}
\usepackage{hyphenat}
\usepackage{xspace}
\usepackage{amsmath}
\usepackage{amsfonts}
\usepackage{hyperref}
\usepackage{url}
\usepackage{booktabs}
\usepackage{multirow}
\usepackage{makecell}
\usepackage{caption}
\usepackage{minibox}
\usepackage{bbm}
\usepackage{bm}
\usepackage{graphicx}
\usepackage{balance}
\usepackage{mathtools}
\usepackage{color}
\usepackage{marvosym}
\usepackage{ifthen}
\usepackage{textcomp}
\usepackage{enumitem}
\usepackage{verbatim}
\usepackage{algorithm}
\usepackage{algorithmic}
\usepackage{numprint}
\usepackage{balance}

\usepackage{amsthm}
\theoremstyle{plain}
\newtheorem{theorem}{Theorem}

\newtheorem{proposition}[theorem]{Proposition}

\newcommand{\chatoDisplayMode}[1]{#1}

\definecolor{MyRed}{rgb}{0.6,0.0,0.0} 
\definecolor{MyBlack}{rgb}{0.1,0.1,0.1} 
\newcommand{\inred}[1]{{\color{MyRed}\sf\textbf{\textsc{#1}}}}
\newcommand{\frameit}[2]{
  \begin{center}
  {\color{MyRed}
  \framebox[.9\columnwidth][l]{
    \begin{minipage}{.85\columnwidth}
    \inred{#1}: {\sf\color{MyBlack}#2}
    \end{minipage}
  }\\
  }
  \end{center}
}

\newcommand{\note}[2][]{\chatoDisplayMode{\def\@tmpsig{#1}\frameit{{\Pointinghand} Note}{#2\ifx \@tmpsig \@empty \else \mbox{ --\em #1}\fi}}}
\newcommand{\abbrevStyle}[1]{#1}
\newcommand{\eg}{\abbrevStyle{e.g.}\xspace}

\newcommand{\Secref}[1]{Sec.~\ref{#1}}

\newcommand{\Tabref}[1]{Table~\ref{#1}}
\newcommand{\Figref}[1]{Fig.~\ref{#1}}

\newcommand{\Appref}[1]{Appendix~\ref{#1}}
\newcommand{\Propref}[1]{Prop.~\ref{#1}}

\newcommand{\xhdr}[1]{\vspace{1.7mm}\noindent{{\bf #1.}}}

\newcommand{\denselist}{ \itemsep -2pt\topsep-10pt\partopsep-10pt }

\newcommand{\textcite}[1]{\citeauthor{#1} \shortcite{#1}}

\newcommand{\cpt}[1]{\textsc{\MakeLowercase{#1}}}

\newcommand{\hide}[1]{}

\makeatletter
\newcommand{\iffont}[2]{\ifthenelse{\equal{\f@family}{#1}}{#2}{}}
\makeatother

  \usepackage{mathptmx}

  \DeclareSymbolFont{greek}{OML}{cmm}{m}{n}
  \DeclareMathSymbol{\alpha}{\mathalpha}{greek}{"0B}
  \DeclareMathSymbol{\beta}{\mathalpha}{greek}{"0C}
  \DeclareMathSymbol{\gamma}{\mathalpha}{greek}{"0D}
  \DeclareMathSymbol{\delta}{\mathalpha}{greek}{"0E}
  \DeclareMathSymbol{\epsilon}{\mathalpha}{greek}{"0F}
  \DeclareMathSymbol{\zeta}{\mathalpha}{greek}{"10}
  \DeclareMathSymbol{\eta}{\mathalpha}{greek}{"11}
  \DeclareMathSymbol{\theta}{\mathalpha}{greek}{"12}
  \DeclareMathSymbol{\iota}{\mathalpha}{greek}{"13}
  \DeclareMathSymbol{\kappa}{\mathalpha}{greek}{"14}
  \DeclareMathSymbol{\lambda}{\mathalpha}{greek}{"15}
  \DeclareMathSymbol{\mu}{\mathalpha}{greek}{"16}
  \DeclareMathSymbol{\nu}{\mathalpha}{greek}{"17}
  \DeclareMathSymbol{\xi}{\mathalpha}{greek}{"18}
  \DeclareMathSymbol{\pi}{\mathalpha}{greek}{"19}
  \DeclareMathSymbol{\rho}{\mathalpha}{greek}{"1A}
  \DeclareMathSymbol{\sigma}{\mathalpha}{greek}{"1B}
  \DeclareMathSymbol{\tau}{\mathalpha}{greek}{"1C}
  \DeclareMathSymbol{\upsilon}{\mathalpha}{greek}{"1D}
  \DeclareMathSymbol{\phi}{\mathalpha}{greek}{"1E}
  \DeclareMathSymbol{\chi}{\mathalpha}{greek}{"1F}
  \DeclareMathSymbol{\psi}{\mathalpha}{greek}{"20}
  \DeclareMathSymbol{\omega}{\mathalpha}{greek}{"21}
  \DeclareMathSymbol{\varepsilon}{\mathalpha}{greek}{"22}
  \DeclareMathSymbol{\vartheta}{\mathalpha}{greek}{"23}
  \DeclareMathSymbol{\varpi}{\mathalpha}{greek}{"24}
  \DeclareMathSymbol{\varrho}{\mathalpha}{greek}{"25}
  \DeclareMathSymbol{\varsigma}{\mathalpha}{greek}{"26}
  \DeclareMathSymbol{\varphi}{\mathalpha}{greek}{"27}
  \DeclareSymbolFont{otone}{OT1}{cmr}{m}{n}
  \DeclareMathSymbol{\Gamma}{\mathalpha}{otone}{0}
  \DeclareMathSymbol{\Delta}{\mathalpha}{otone}{1}
  \DeclareMathSymbol{\Theta}{\mathalpha}{otone}{2}
  \DeclareMathSymbol{\Lambda}{\mathalpha}{otone}{3}
  \DeclareMathSymbol{\Xi}{\mathalpha}{otone}{4}
  \DeclareMathSymbol{\Pi}{\mathalpha}{otone}{5}
  \DeclareMathSymbol{\Sigma}{\mathalpha}{otone}{6}
  \DeclareMathSymbol{\Upsilon}{\mathalpha}{otone}{7}
  \DeclareMathSymbol{\Phi}{\mathalpha}{otone}{8}
  \DeclareMathSymbol{\Psi}{\mathalpha}{otone}{9}
  \DeclareMathSymbol{\Omega}{\mathalpha}{otone}{10}
  \DeclareSymbolFont{syms}{OML}{cmm}{m}{it}
  \DeclareMathSymbol{\partial}{\mathord}{syms}{"40}
  \DeclareMathAlphabet{\mathbold}{OML}{cmm}{b}{it}
  \DeclareSymbolFont{largesymbols}{OMX}{cmex}{m}{n}

\usepackage{microtype}
\usepackage{multirow}
\usepackage[pass]{geometry}

\usepackage{graphicx}  %Required
\usepackage{subcaption}

\usepackage{svg}

\usepackage{amsmath}
\usepackage{amssymb}
\usepackage{booktabs}
\usepackage{bbm}
\usepackage{amsfonts}       % blackboard math symbols
\usepackage{nicefrac}       % compact symbols for 1/2, etc.
\usepackage{microtype}      % microtypography
\usepackage{amsthm}
\usepackage{thmtools, thm-restate}
\usepackage{natbib}
\usepackage{hyphenat}
\usepackage{xcolor}

\usepackage{mathtools}

\aclfinalcopy % Uncomment this line for the final submission
\title{Better than Average: Paired Evaluation of NLP Systems}

\author{Maxime Peyrard$^*$, Wei Zhao$^\dagger$,
Steffen Eger$^\dagger$, Robert West$^*$\\
    $^*$EPFL, Switzerland \\
    $^\dagger$Technische Universit\"at Darmstadt, Germany \\
    {\tt \{maxime.peyrard,robert.west\}@epfl.ch} \\
    {\tt \{zhao,eger\}@aiphes.tu-darmstadt.de}\\
    
  }

\date{}

\begin{document}
\maketitle

\begin{abstract}
Evaluation in NLP is usually done by comparing the scores of competing systems independently averaged over a common set of test instances.
In this work, we question the use of averages for aggregating evaluation scores into a final number used to decide which system is best, since the average, as well as alternatives such as the median, ignores the pairing arising from the fact that systems are evaluated on the same test instances. 
% The scores of competing systems are thus by test instance. 
We illustrate the importance of taking the instance\hyp level pairing of evaluation scores into account and demonstrate, both theoretically and empirically, the advantages of aggregation methods based on pairwise comparisons, such as the Bradley--Terry (BT) model, a mechanism based on the estimated probability that a given system scores better than another on the test set. By re-evaluating 296 real NLP evaluation setups across four tasks and 18 evaluation metrics, we show that the choice of aggregation mechanism matters and yields different conclusions as to which systems are state of the art in about 30\% of the setups.
To facilitate the adoption of pairwise evaluation, we release a practical tool for performing the full analysis of evaluation scores with the mean, median, BT, and two variants of BT (Elo and TrueSkill), alongside functionality for appropriate statistical testing.
\end{abstract}

\section{Introduction}

Research is driven by evaluation results, with attention and resources being focused on methods identified as state of the art (SotA). The proper design of evaluation methodology %ies 
 is thus crucial to ensure %the 
progress %of 
in 
the field.
% Designing proper evaluation methodologies is critical because the evaluation results guide research by promoting the specific methods identified as state-of-the-art.
In NLP, %the 
evaluation %is usually done by 
usually consists in 
comparing the averaged scores of competing systems over a common set of test instances. Indeed, averaging scores independently for each system and declaring the one with the highest average to be best is particularly simple, well understood, and mirrors the expected risk minimization paradigm used to train systems.

Here, we %propose a critical assessment of 
critically assess 
the specific choice of the average to aggregate evaluation scores. 
In particular, we emphasize %the natural
that there is a natural
\emph{instance\hyp level pairing} between the evaluation scores of systems, which
aggregation mechanisms such as the mean or median fail to take into account: 
%
%, and, thus, 
%fail to %distinguish systems that, for example, have the same set of scores but paired (ordered) differently.
%account for: 
as they produce a score for each system independently, systems that have the same set of scores (but potentially in different order) cannot be distinguished. 

%In particular, we emphasize the natural
%\emph{instance-level pairing} between the evaluation scores of systems. 
%Aggregation mechanisms like the mean or the median produce a score for each system independently, and, thus, fail to distinguish systems that, for example, have the same set of scores but paired (ordered) differently.

\begin{figure}[t]
    \centering
    \includegraphics[width=0.8\columnwidth]{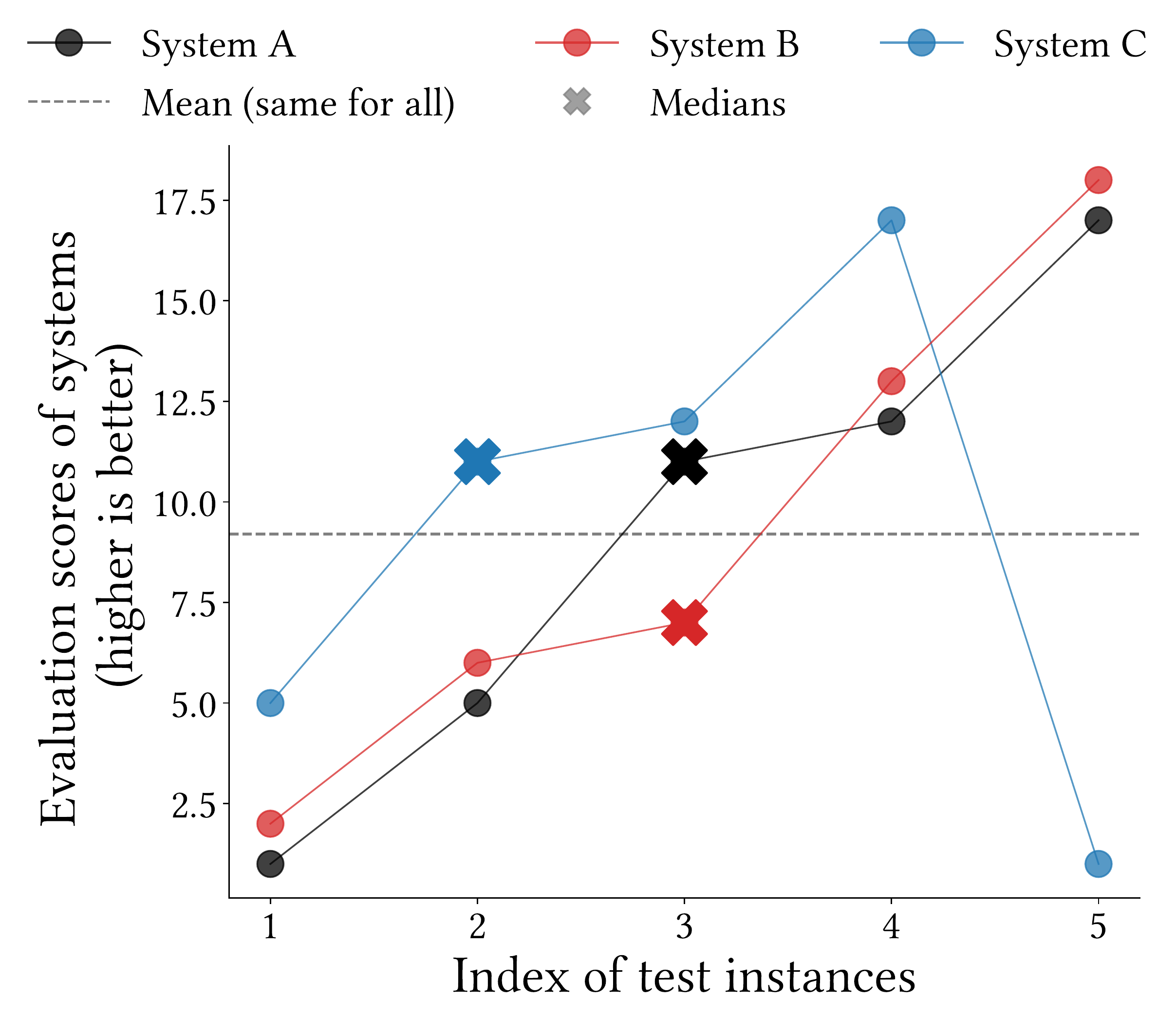}
    \caption{\textbf{Motivating example} (synthetic data). Evaluation scores of systems $A$, $B$, and $C$ for five test instances. All systems have the same mean.
    % , and their medians are represented by $\times$ markers. 
    $C$ is better than $A$ on all instances but one, so BT declares $C>A$
    % , whereas the medians and means of $A$ and $C$ are the same. 
    Also, $B$ is better than $A$ on all instances but one, so BT declares $B>A$, whereas the median of $A$ is greater, and the means are the same.
    Overall, mean and median fail to capture the complex instance-level pairing.
    % There is a complex instance-level pairing of systems that the mean and median fail to capture.
    % \todo[inline]{Wei: One mistake: the Medians of A, B are NOT the same. }
    }
    \label{fig:figure_1}
\end{figure}

Consider the three systems $A$, $B$, and $C$ compared on five test instances in \Figref{fig:figure_1}. Despite a complex pairing structure, they all have the same mean score across test instances. Moreover, even though $B$ is better than $A$ on all test instances but one, the median of $A$ is greater than the median of $B$. 
% \todo{Wei: Maybe consider to drop this paragraph (or merge into the caption) so as to avoid repeating the content in the caption.}

In this work, 
we discuss an alternative aggregation mechanism: the Bradley--Terry (BT) model \cite{bt_one}.
% \todo{SE: We discuss only this one, so I would omit "like the"} 
%\todo{Wei: Consider to briefly mention Median to stay consistent with Fig.~1.}
BT compares systems for each test instance and estimates the latent strength of systems based on how frequently one system scores higher than another.
Such paired mechanisms have already been successfully used to aggregate human judgments \cite{novikova-etal-2018-rankme, sedoc-ungar-2020-item}; for example, WMT evaluation protocols regularly employ TrueSkill \cite{true_skill}, a Bayesian variant of BT \cite{sakaguchi-etal-2014-efficient}. 
% \todo{Wei: Maybe avoid saying WMT workshops. We could say WMT evaluation protocol employs TrueSkill to aggregate human judgments.}
% \todo[inline]{SE: But WMT is not human evaluation ? If WMT uses this as a mechanism for comparing and ranking MT systems, this would be decrease the novelty of the contribution by a lot, IMO}

\xhdr{Contributions}
We contribute the first comprehensive analysis of the BT model (especially vis-\`a-vis mean and median) 
as an aggregation mechanism for comparing system scores in NLP.

\textbf{(i)} 
We illustrate the importance of accounting for instance-level pairing and discuss the conditions under which the mean, median, and BT disagree about the ordering of systems. In \Secref{sec:theory}, we draw parallels with the field of statistical testing, where \emph{paired statistical tests} are recommended when comparing paired variables. 
Thus, we argue that paired aggregation mechanisms such as BT are more robust alternatives to the mean and median. %demonstrate it 
We support this argument with simulations in \Secref{sec:simulations}.

% For any pair of systems, BT declares as best the system that scores higher on the majority of instances. A researcher believing that this criterion would lead to inconsistent ordering of systems should already be skeptical about the evaluation setup: either the test set is not representative and/or the evaluation metrics is not appropriate.

\textbf{(ii)} 
We show that the differences between mean, median, and BT matter in practice. By re\hyp evaluating $296$ real NLP evaluation setups across four tasks and 18 evaluation metrics, different aggregation mechanisms yield different conclusions as to which systems are SotA in about 30\% of the setups (\Secref{sec:experiments}). These results hold when replacing BT by the Elo \cite{elo1978rating} and TrueSkill variants.

\textbf{(iii)} %In \Secref{sec:discussion}, 
We discuss further advantages and potential limitations of BT, alongside possible resolutions, in \Secref{sec:discussion}.

\textbf{(iv)} 
We recommend replacing the mean by BT in future evaluations of NLP systems. To ease the adoption of more robust aggregation mechanisms, we release \textit{Pairformance},%
\footnote{\url{https://github.com/epfl-dlab/pairformance}} a practical tool for performing full analyses of evaluation scores with mean, median, BT, and two variants of BT (Elo and TrueSkill). The tool reports paired evaluation results alongside appropriate statistical testing for all five aggregation mechanisms and various visualization functionalities to elucidate the pairing structure between system scores.

Code and data for replicating our analyses and experiments is available online.%
\footnote{\url{https://github.com/epfl-dlab/BT-eval}}

\section{Aggregation of evaluation results}
\label{sec:framework}
In this section, we briefly present the three aggregation mechanisms we consider.
% \todo{Wei: maybe rewrite this sentence}

\subsection{Terminology}
A standard evaluation setup typically consists of four elements:
\begin{enumerate}
\denselist
    \item At least two \textbf{systems,} $A$ and $B$, to compare, with latent strengths $\lambda_A$ and $\lambda_B$ that we aim to estimate.
    
    \item A \textbf{test set} $T=\bigl\{(x_l,y_l)\: :\: l=1,\ldots,n\bigr\}$ consisting of $n$ test instances, where $x_l$ is the input and $y_l$ is the ground-truth target output.
    
    \item An \textbf{evaluation metric} $M$ for scoring system outputs based on target outputs $y_l$, resulting in the sequence of \textbf{evaluation scores} $\mathcal{M}_{A} = \langle M(A(x_l), y_l):\: l=1,\ldots,n\rangle$ for system $A$.
    
    \item An \textbf{aggregation mechanism} $\Theta$ that decides whether system $A$ is better than $B$ based on the evaluation scores of the two systems. 
    We use $\Theta_{T,M}(A, B)
        = \Theta(\mathcal{M}_{A}, \mathcal{M}_{B})$ 
    to denote the comparison mechanism between $A$ and $B$ on the test set $T$ with evaluation metric $M$. 
    Here, $\Theta$~outputs its guess about which system is the best (or declares the comparison inconclusive if the difference is not statistically significant). For simplicity, we drop the dependency on $T$ and $M$ in the notation, simply writing $\Theta(A,B)$.
    % We now drop the explicit dependence on $T$ and $M$ as it is not ambiguous and simply note $\Theta(A,B)$. \todo{Rewrite to "We drop $T$ and $M$ for simplicity since $\Theta$ is oftentimes shared over tasks and metrics"?}
\end{enumerate}

For example in text summarization, $x_l$ is a source document from the test set, $y_l$ its corresponding reference summary, and $M$ might be \cpt{ROUGE} \cite{Lin:2004}. The decision mechanism $\Theta$ usually compares the individual systems' mean evaluation scores, where the system with the highest mean score (here mean \cpt{ROUGE} score) is declared better.

% \begin{equation}
% \label{eq:mean_def}
% \begin{split}
%     &\Theta(S, B) \\
%     &= \begin{cases}
%     S, \ \text{if }\sum\limits_{l} M(A(x_l), y_l) >^{*} \sum\limits_{l} M(B(x_l), y_l)\\
% B, \ \text{if } \sum\limits_{l} M(B(x_l), y_l) >^{*} \sum\limits_{l} M(A(x_l), y_l)\\
% \text{inconclusive}, \ \text{otherwise}.
% \end{cases}
% \end{split}
% \end{equation}

% We use $>^*$ to denote `statistically significantly greater' according to appropriate test. 
% Then, $\Theta$ decides that $A$ is better than $B$ if the average score of $A$ is better than the average score of $B$.\footnote{Potentially does not conclude if the different is not significant.}

\xhdr{Consistent evaluation result} We say that the outcome of such an evaluation is \emph{consistent} if it recovers the ordering of systems implied by the inherent strengths of systems: %\todo{SE: do we make this notation up or is it standard terminology? The inherent strengths ($\lambda_S,\lambda_B$) have not been introduced but it is probably still clear here, I would say}
% \begin{equation}
    $\Theta(A, B) = A \iff \lambda_A > \lambda_B$.
% \end{equation}

\xhdr{Probabilistic model}
As commonly done in the literature on statistical testing, we view the evaluation scores of a system $A$ as $n$ indexed random variables: $X_A^{(l)}, \: l=1,\ldots,n$, where $n$ is the size of the test set. Note that this sequence of random variables is not necessarily i.i.d. 
% \todo{Wei: why use $n$ RVs instead of one RV with $n$ observations? MP: I think (a) it is a simpler way to really show the pairing because the variables are paired by the indices, (b) this is the notations used by statistical tests like the sign test, (c) With n rvs, they can have different distributions, the i.i.d. case being very specific case where the pairing would not matter. }
% Similarly, the evaluation scores of system $B$ are noted $X_B^{(l)}$.
% We note $X_T$ the random variable from which test instances $(x, y)$ are drawn: $X_T \sim \mathbb{P}_{T}$.
% Similarly, we note $X_A$ the random variable that represents the evaluation scores of system $A$ and $X_B$ for system $B$.
% These random variables depend on the choice of the metric but also on the test samples as illustrated by the graphical model in.
% To obtain the evaluation scores of systems $A$ and $B$, it goes as follows:
% \begin{enumerate}
%     \item Sample a test instance $X_T = (x,y)$ from $\mathbb{P}_{T}$
%     \item Compute the scores of the systems $X_k|(x,y) = M(k(x), y), k \in \{A,B\}$
% \end{enumerate}
Furthermore, even though systems $A$ and $B$ are independent, their evaluation scores are not, since there is an instance-level \textbf{pairing.}
%because the systems are evaluated on the same test instances: $X_A^{(l)}$ is paired with $X_B^{(l)}$. 
Intuitively, knowing the score of $A$ on an instance $(x_l, y_l)$ can provide information about the expected performance of $B$. For example, if $A$ scores highly because $(x_l, y_l)$ is an easy instance, one might expect $B$ to also score highly. 
% \todo{Wei: what properties? Maybe rewrite to "one could query on which instance (or how many instances) system A performs better than system B"}

\subsection{Aggregation mechanisms}

We now introduce three aggregation mechanisms~$\Theta$. We investigate their properties in subsequent sections.

\xhdr{Mean}
This is the current standard: the system with the highest average score is declared the strongest. We denote this aggregation mechanism as \cpt{MEAN}.
% the arithmetic average of the scores over test instances is the estimated strength of the system. 
The average score of system $A$ is computed as $E_{A} = \frac{1}{n} \sum\limits_{l=1}^n X_A^{(l)}$.
% In terms of random variables, the average of the evaluation scores of system $A$ corresponds to integrating out the dependence on $X_T$:
% \begin{align}
%     E_{A} = \frac{1}{n} \sum\limits_{l=1}^n X_A^{(l)},
% \end{align}
% where $x_A^{(k)}$ is the observed score of system $A$ on the $k$-th instance.

% where the second line show the sample estimate of the first line.

% It has been demonstrated already in Eq.~\eqref{eq:mean_def}.

\xhdr{Median}
The median is an interesting alternative to the mean because it is robust to outliers.
% An interesting alternative to the mean is the median. The median, contrary to the mean, is robust to outliers.
% In our case, it is straightforward to replace the mean by the median. 
Here, the system with the highest median score is declared to be the strongest. 
The median score $M_A$ of a system $A$ is the central value in the sorted list of evaluation scores of $A$.
% of $x_A^{(l)}$ such that $50\%$ of test instances  and it is the $\frac{n+1}{2}$ largest number in the list of scores of $A$ if $n$ is odd and the average between the $\frac{n}{2}$ and $\frac{n+1}{2}$ largest numbers if the $n$ is even.
% \begin{align}
%     &\mathbb{M}_{l}\left[X_A^{(l)}\right] \\
%     &\frac{n+1}{2}^{th} \text{ element in the sorted list of scores } A,
% \end{align}
% where the second line is the sample median.
We denote this aggregation mechanism as \cpt{MEDIAN}. 

\xhdr{Bradley-Terry}
The third option examined here is the Bradley--Terry (BT) model \cite{bt_one}. While \cpt{Mean} and \cpt{Median} compute scores for systems $A$ and $B$ independently, BT is a function of the joint random variable $\left(X_A^{(l)},X_B^{(l)}\right)$.
BT estimates the relative strengths $\hat{\lambda}_A$ and $\hat{\lambda}_B$ of the two systems $A$ and $B$, by comparing the evaluation scores for each test instance: 
\begin{equation}
    \mathbb{P}(A > B) = \frac{\hat{\lambda}_A}{\hat{\lambda}_A + \hat{\lambda}_B}.
\end{equation}
Intuitively, $\mathbb{P}(A > B)$ is the probability that, for any given test instance, $A$ scores higher than $B$. The BT model chooses $\hat{\lambda}_A$ and $\hat{\lambda}_B$ in order to best explain the observations. The system with the highest $\hat{\lambda}$ is declared strongest.

When considering only two systems, the latent strength $\hat{\lambda}_A$ is the number of instances for which $A$ scores better than $B$ (and similarly for $\hat{\lambda}_B$). 
When the number of systems is greater than two, BT solves an iterative optimization algorithm that is guaranteed to converge to a unique solution \cite{bt_one}. We give details about BT and its computation in the general case in \Appref{app:bt_details}.

We denote as \cpt{BT} the decision mechanism based on the BT model. While it is much less common than \cpt{MEAN} and \cpt{Median}, we will see below that \cpt{BT} satisfies interesting properties making it a more robust alternative. 

\section{Comparison of assumptions}
\label{sec:theory}
Since the roles played by $A$ and $B$ are symmetrical, we now assume without loss of generality that system $A$ is better, i.e., $\lambda_A > \lambda_B$.  

\begin{proposition}
\label{prop:main_prop}
If $\lambda_A > \lambda_B$ then
% and the evaluation scores of $A$ and $B$ are represented by the random variables $X_A$ and $X_B$. 
\begin{itemize}
\denselist
    \item \cpt{Mean} consistent $\iff  E_A -  E_B > 0$,
    \item \cpt{Median} consistent $\iff M_A - M_B > 0$,
    \item \cpt{BT} consistent $\iff M_{A-B} > 0$,
\end{itemize}
where $E_S$ and $M_S$ are the mean and median of the evaluation scores of system $S$, and 
$M_{A-B}$ is the median of the differences between the evaluation scores of $A$ and $B$. Note that $E_S, M_S$, and $M_{A-B}$ are all random variables.
% random variables obtained from taking the mean and median of the list of random variables from the evaluation scores of system $S$. 
% \todo{Wei: For simplicity, maybe rewrite the first sentence to '$E_S$, and $M_S$ are the mean and median of the evaluation scores of system $S$.'}
% $X_S - X_B$ is a random variable obtained by first sampling a test instance $(x_l, y_l)$ and then computing the difference in evaluation score between $S$ and $B$. It reflects the particular paired structure of the evaluation.
\end{proposition}
The proof is given in \Appref{sec:proof}.
% (for \cpt{MEAN} and \cpt{MEdian} the statement is direct). 
% \todo{Wei: What is the logic of the proof in Appendix B, e.g., to prove the "iff" statement?} 
% (note that for \cpt{mean} and \cpt{median} the statement is direct given the definition).
% \begin{proof}
% The case of \cpt{MEAN} and \cpt{MEDIAN} are direct. 
% For \cpt{BT}, observe that it correctly gives $A$ better than $B$ iff $\mathbb{P}(X_A > X_B) > \mathbb{P}(X_B > X_A) \iff \mathbb{P}(X_A > X_B) > \frac{1}{2} \iff \mathbb{P}(X_A - X_B > 0) > \frac{1}{2}$, for more than $50\%$ of instances $X_A^{(l)} - X_B^{(l)}> 0$ and thus the median is positive.
% \end{proof}
Note that, whereas the expectation is linear ($E_A - E_B = E_{A-B}$), the median is not (in general, $M_A - M_B \neq M_{A-B}$).
% While the expectation is linear $\mathbb{E}[X_A^{(l)} - X_B^{(l)}] = \mathbb{E}[X_A^{(l)}] - \mathbb{E}[X_B^{(l)}]$, the median is not: $\mathbb{M}[X_A^{(l)} - X_B^{(l)}] \neq \mathbb{M}[X_A^{(l)}] - \mathbb{M}[X_B^{(l)}]$. 

% \paragraph{Why not use directly the median of differences as the aggregation mechanism?} \todo{Wei: the difference of medians? $M_A - M_B$ or $M_{A-B}$?}
% It would be possible and certainly useful for comparing systems under paired data and potential outliers. However, the median of differences does not output one score for each systems (see the discussion about the sign test in \newcite{dror-etal-2018-hitchhikers}). \cpt{BT} produces the $\hat{\lambda}$ estimates for each system that are overall consistent with this statistic, as shown by \Propref{prop:main_prop}.

% The non-linearity of the median is the source of what makes the BT and median methods differ.

% \xhdr{Relation between \cpt{mean} and \cpt{median}}
% It is widely known that the median is an alternative to the mean robust to outliers. This can be seen here because if the evaluation scores of both systems $X_A|X_T$ and $X_B|X_T$ are non-skewed, $\mathbb{E}[X_A|X_T] = \mathbb{M}[X_A|X_T]$ and $\mathbb{E}[X_B|X_T] = \mathbb{M}[X_B|X_T]$, \cpt{MEAN} and \cpt{Median} produce the same ordering of systems. Otherwise, the presence of outliers may can switch the decision of \cpt{MEAN}.

\xhdr{Robustness to ouliers}
The mean is not robust to outliers:
% $\mathbb{E}_{l}\left[X_A^{(l)} - X_B^{(l)} \right]$ 
$E_{A-B}$
can be swayed above or below the threshold of $0$ by a small number of test instances for which the difference between system scores is large. 
% The median offers a robust alternative to the mean 
On the contrary, the median is a robust statistic that cannot be easily influenced by outliers.
% $\mathbb{M}_{l}\left[X_A^{(l)}\right] - \mathbb{M}_{l} \left[X_B^{(l)} \right]$ cannot be swayed above or below the $0$ threshold just by the presence of outliers.
Similarly, \cpt{BT} is robust to outliers because its decision is based on the median of differences~$M_{A-B}$.
% , it cannot be swayed above or below the $0$ threshold by few outlier instances. 
% \todo{Wei: it seems the sentence after the last ',' is unnecessary.}

\xhdr{Importance of pairing}
The critical difference between \cpt{BT}, \cpt{Mean}, and \cpt{Median}, is that only \cpt{BT} preserves the pairing information. Both \cpt{MEan} and \cpt{Median} compute a statistic from the (unordered) set of scores $X_A^{(l)}$ and $X_B^{(l)}$ independently and then compare the aggregate statistics, losing the pairing structure.
% The pairing is lost and ignored in the process. 
% Indeed, any permutation of the index $\sigma(l)$ leaves the mean and median unchanged but greatly changes the pairing.
If the pairing actually does not matter, any permutation of the indices of system scores leaves the distribution of paired evaluation scores unchanged.
% , i.e. $\forall \sigma, \forall l, (X_{A}^{(\sigma(l))}, X_B^{(l)})$ and $(X_{A}^{(l)}, X_B^{(l)})$ have the same distribution.
This happens, for example, when both $X_A^{(l)}$ and $X_B^{(l)}$ are i.i.d.\footnote{More generally, when the two sequences  of random variables are exchangeable.}

However, in the general case, the pairing matters. One particular example is when there exist different types of test instances and systems behave differently for different types,
% \todo{Wei: Shorten to "there exists different types of test instances on which systems behave differently"?} 
e.g., when there are \emph{easy} instances on which all systems have higher scores.
% , or instances for which some systems have been crafted for and thus performs particularly well.
For example, consider the three systems and their evaluation scores on five test instances in \Figref{fig:figure_1}.
% in \Tabref{tab:example_median}.
System $A$ is worse than $C$ on all instances but one, so $C > A$ according to \cpt{BT}, yet the median of $A$ is greater than the median of $C$ (10 vs.\ 7). At the same time, $B$ outperforms $C$ on all instances but one, so $B > C$ according to \cpt{BT}. 
% Yet, $A$ and $B$ have the exact same set of scores but in a different order, i.e., different pairing.
For \cpt{Median} and \cpt{mean}, which ignore the pairing, $A$ and $B$ are completely equivalent, even though there is a clear difference regarding which system is more likely to be the best. This difference is revealed in the pairing structure. In general, any mechanism ignoring the pairing cannot capture the difference between $A$ and $B$.

% Now, system $B$ is better than $C$ on all instances but one, and $B > C$ according to \cpt{BT} which now agrees with \cpt{Median}. Therefore, \cpt{Median} which ignores the pairing cannot capture the difference between system $A$ and $B$ in comparison to $C$. Similarly, for the \cpt{Mean} system $A$ and $B$ are perfectly equivalent. In general, any mechanism that ignores the pairing cannot capture the differences between $A$ and $B$.

\xhdr{Choosing an aggregation mechanism}
In \Propref{prop:main_prop}, we stated the conditions for each mechanism to be \emph{consistent}. Choosing an aggregation mechanism for a specific evaluation setup boils down to deciding what condition is more likely to hold in the setup. Note that none of the conditions implies any other condition in \Propref{prop:main_prop}.

When comparing \cpt{BT} against \cpt{MEAN} (or \cpt{MEdian}), there are three possible scenarios: (i) \cpt{BT} agrees with \cpt{Mean} (or \cpt{median}), (ii) \cpt{BT} is consistent but \cpt{mean} (or \cpt{median}) is not, and (iii) \cpt{mean} (or \cpt{median}) is consistent but \cpt{BT} is not. 
% \todo{Wei: Prop. 1. indicates that MEAN, Median and BT consider to be consistent with $\lambda_A \ge \lambda_B$ under certain conditions (e.g., $E[X_A-X_B] \ge 0$), but now the conditions (e.g., outliers) under which they are inconsistent are discussed. Did I follow the logic correctly?}
% \begin{enumerate}
% \denselist
%     \item[(i)] \cpt{BT} agrees with \cpt{Mean} (or \cpt{median}).
%     \item[(ii)] \cpt{BT} is consistent but \cpt{mean} (or \cpt{median}) is not. 
%     \item[(iii)] \cpt{mean} (or \cpt{median}) is consistent but \cpt{BT} is not. 
%     % They also disagree on the ordering of systems.
    
% \end{enumerate}

In case (i), it does not matter whether we use \cpt{BT} or \cpt{MEAN} (or \cpt{Median}).
% \todo{Wei: Maybe briefly mention the score distribution in this scenario, similar to the other two cases. }

In case (ii), for most instances, the better system has a higher score than the worse system, but \cpt{Mean} (or \cpt{median}) fails. For example, \cpt{Mean} may be swayed by outliers, and \cpt{Median} may be swayed by jumps in score lists as in the example above.
% the mean of the best system is pulled down by few outliers and/or the mean of the worst system is pulled up by few outliers. Similarly, the \cpt{median} can fail because of the global structure of the pairing as in the example above.

In case (iii), for most instances, the better system has a lower score than the worse system, yet particular variations in the marginals make the \cpt{mean} or \cpt{Median} get the ordering correct. This is a very peculiar scenario: for \cpt{mean}, it implies that on the few instances on which the better system did better, the difference between evaluation scores was large enough to lift the mean of the better system above the other. We argue that if one really believes that the evaluation setup is likely to be in case~(iii), then one does not trust the evaluation setup in the first place. It corresponds to assuming that the observed scores are inconsistent for the majority of test instances. If this is the case, one should rather improve the evaluation setup (\eg, metric, test set) in order to be more representative of the phenomena that one desires to capture.

Overall, the condition making \cpt{BT} consistent appears to be the most natural one. Trusting \cpt{mean} or \cpt{median} more than \cpt{BT} implies holding an unintuitive belief about the evaluation setup, namely that the better system does worse than the worse system on a majority of test instances. 

From another perspective, the random variables $E_{A} - E_{B}$ (\cpt{mean}) and $M_A - M_B$ (\cpt{median}) are less likely to be (correctly) greater than zero in the presence of (i)~complex pairing structures or (ii)~outliers. The variable $M_{A-B}$ (\cpt{BT}), on the contrary, is not affected by complex pairings or outliers. 
% However, like \cpt{MEAN} and \cpt{Median}, \cpt{BT} remain influenced by the noise and data imbalance

% If one has reason to believe that about the evaluation setup, one should probably first re-assess and improve the evaluation setup (metric, test set) to be more representative of the phenomena that one desires to capture.

% Overall, it seems that the assumption underlying \cpt{BT} is more natural than the assumptions underlying \cpt{mean} or \cpt{median}. Indeed, preferring \cpt{mean} or \cpt{median} to \cpt{BT} implies believing that, for most test instances, the best system is actually worst than the worst system. As long as one believes that for most instances, the best system has the better score, then \cpt{BT} is trustworthy.
% Believing that should makes us skeptical about the evaluation setup.
 
% \begin{table}
%         \small
%         \centering
%         \resizebox{0.75\columnwidth}{!}{
%         \begin{tabular}{l|ccccc}
%         \toprule

%          & test 1 & test 2 & test 3 & test 4 & test 5 \\
%         \midrule
%         $A$     & 1 & 5 & 10 & 12 & 17 \\
%         $B$     & 5 & 10 & 12 & 17 & 1 \\
%         $C$     & 2 & 6 & 7 & 13 & 18 \\
%         \bottomrule
%         \end{tabular}
%         }
%         \caption{Example of $3$ systems evaluation scores.}
%         \label{tab:example_median}
% \end{table}

\begin{figure*}
    \centering
    \begin{subfigure}[t]{0.30\textwidth}
    \centering
        \includegraphics[width=\linewidth]{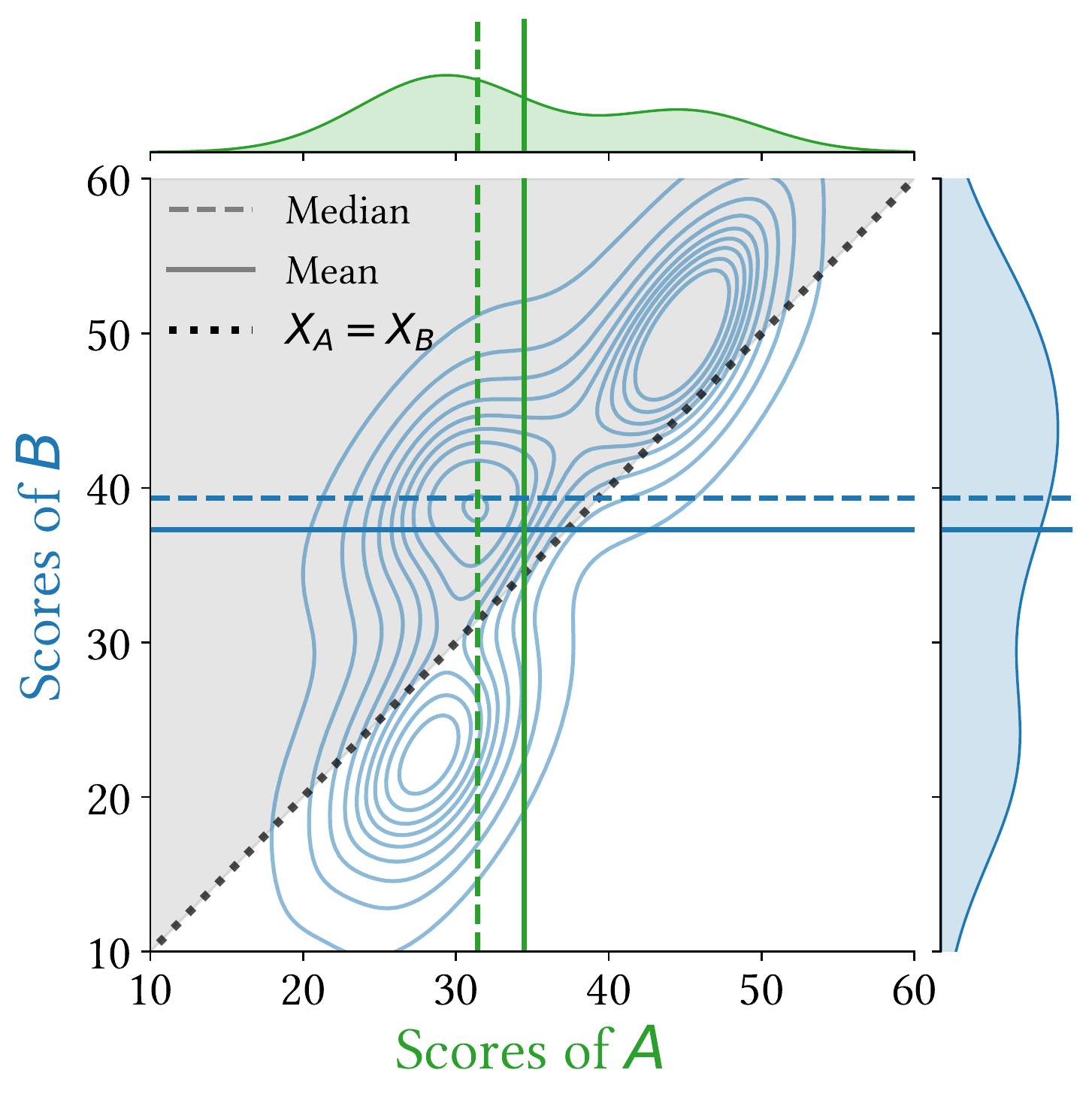}
        % \caption{permuation a}
        % \label{fig:geom_perm_a}
    \end{subfigure}%
    \begin{subfigure}[t]{0.30\textwidth}
    \centering
        \includegraphics[width=\linewidth]{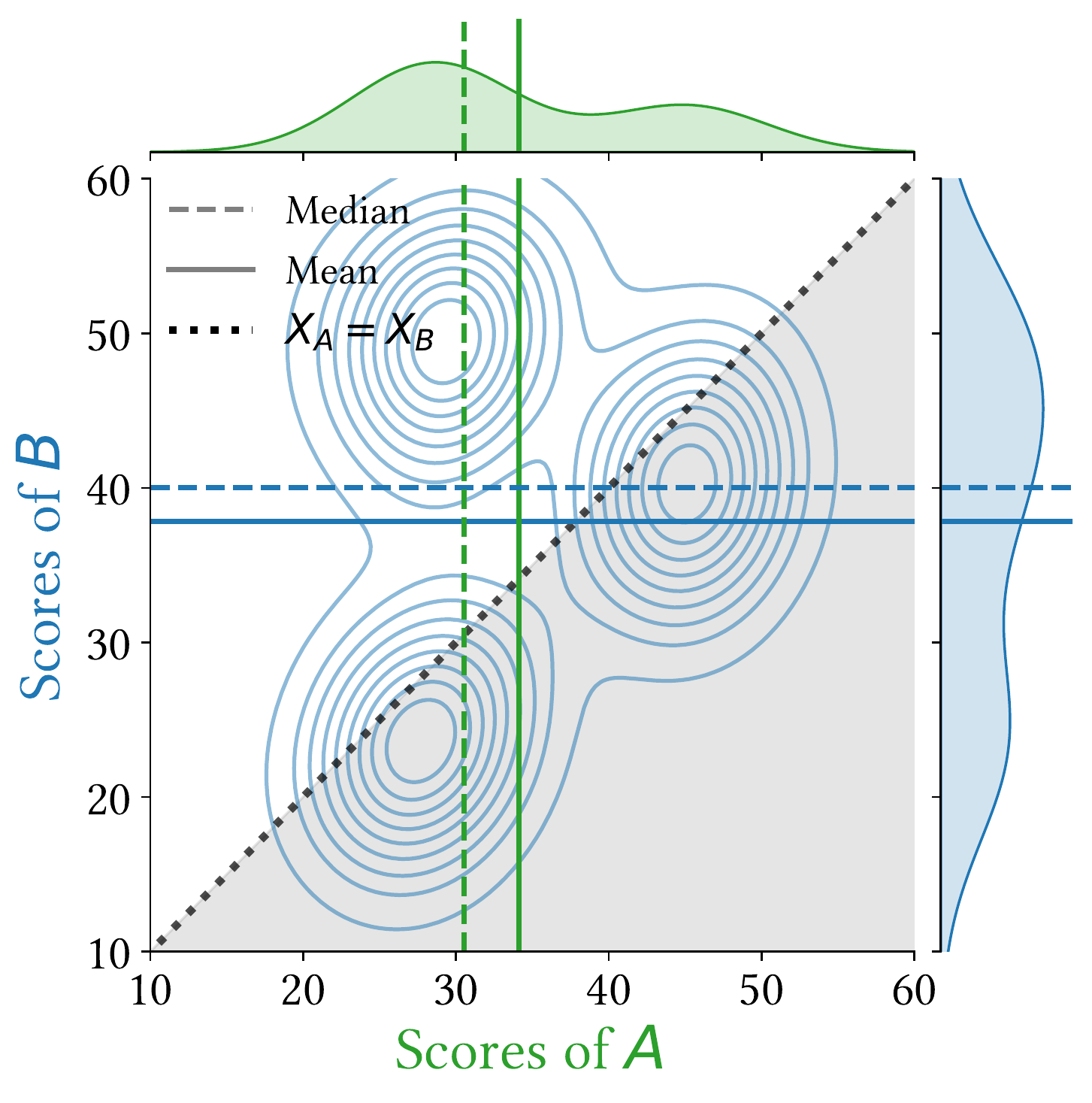}
        % \caption{permuation b}
        % \label{fig:geom_perm_b}
    \end{subfigure}
    \begin{subfigure}[t]{0.30\textwidth}
    \centering
        \includegraphics[width=\linewidth]{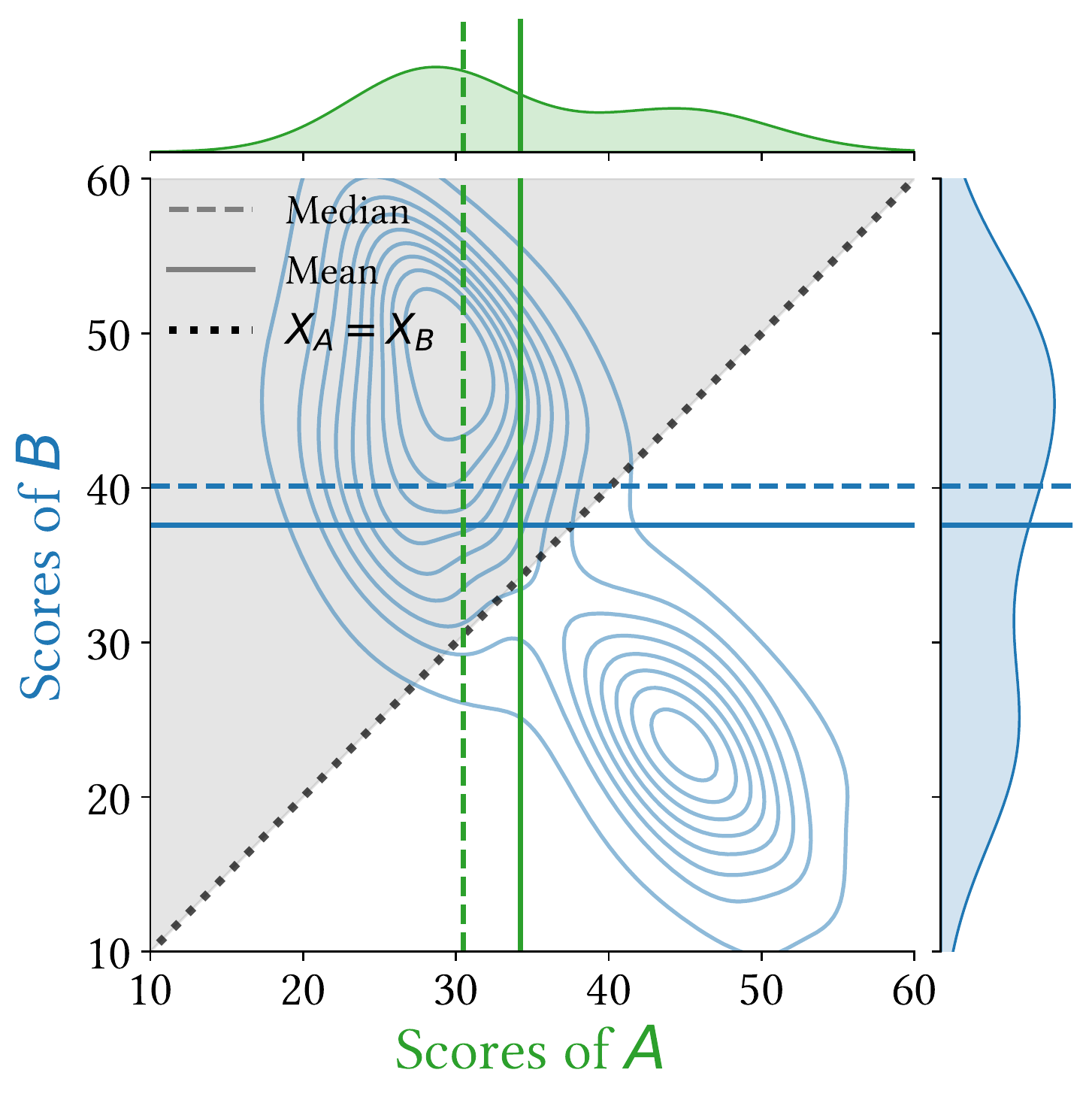}
        % \caption{permuation c}
        % \label{fig:geom_perm_c}
    \end{subfigure}
    % \begin{subfigure}[t]{0.24\textwidth}
    % \centering
    %     \includegraphics[width=\linewidth]{images/pairs_perm_4.pdf}
    %     \caption{permuation d}
    %     \label{fig:geom_perm_d}
    % \end{subfigure}
    \caption{These 2D plots represent the distribution of test instances with coordinates given by the scores of the two systems being compared, i.e., the $x$-axis is the score $X_A^{(l)}$ of system $A$ on some test instance $(x_l, y_l)$, and the $y$-axis is the score $X_B^{(l)}$ of system $B$ on the same instance. While the $3$ plots represent different instance-level performances of $A$ and $B$, the marginal (unpaired) distribution of scores of $A$ and $B$ remain unchanged. From such 2D plots, not only do we see the global structure of the pairing between the scores of $A$ and $B$, we can also read off the decision of \cpt{mean}, \cpt{median} and \cpt{BT} based on simple \textbf{geometrical criteria:} (i) if the prolongation of the means intersect above the $X_A=X_B$ line, then \cpt{Mean} predicts that $B$ is better, (ii) if the prolongation of the medians intersect above the $X_A=X_B$ line, then \cpt{Median} predicts that $B$ is better, (iii) if there is more mass in the upper-left triangle, then \cpt{BT} predicts that system $B$ is better. The latter case corresponds to most of the test instances being located in the upper-left triangle ($B>A$). The half-space with more mass is shaded. 
    % We see that in the middle plot \cpt{BT} disagrees with \cpt{Mean} and \cpt{median} as it predicts $B$ is the best. \cpt{Mean} and \cpt{Median} cannot capture the difference between these different setups. 
    }
    \label{fig:geom_crit}
\end{figure*}

\subsection{Graphical criterion}
\Figref{fig:geom_crit} summarizes the problem of ignoring the pairing and offers a graphical criterion to understand the decisions made by \cpt{Mean}, \cpt{median}, and \cpt{BT}.
In each plot, the densities are estimated by placing test instances at coordinates given by the evaluation scores of the two systems. The evaluation scores of $A$ (green) are on the $x$-axis, and the evaluation scores of $B$ (blue) on the $y$-axis. 
We also plot the \emph{marginal distributions} of evaluation scores, from which we can read off means and medians. When the mean of $X_B^{(l)}$ is greater than that of $X_A^{(l)}$, the two extended lines representing the means meet in the upper triangle (above the line $X_A = X_B$), and analogously for the median.
But mean and median being only functions of the marginals, they completely ignore the pairing. \Figref{fig:geom_crit} illustrates this by depicting three completely different pairing structures where the marginals (and thus the means and medians) of $A$ and $B$ remain unchanged. (In \Appref{sec:paring-examples}, we explain how to generate infinitely many such examples.)
% Thus, the mean and median of $A$ and $B$ are also exactly the same in these three figures.
% In the figure, this is illustrated by having 4 different pairing (i.e., different structure of the evaluation) that still give rise to the exact same two marginal evaluation score distributions. 
% \cpt{Mean}, \cpt{Median} and any other function of the evaluation score distribution that ignores the pairing cannot capture such nuances in the evaluation setup.
On the contrary, \cpt{BT}, being a property of the pairing (the 2D density), predicts that $B$ is better than $A$ when there is more mass in the upper triangle, i.e., more instances for which $B$ scores higher than $A$.
In the middle figure, the pairing indicates that $A$ is better than $B$, in disagreement with the decisions of \cpt{MEAN} and \cpt{Median}.
% We discuss how to generate infinitely many such examples in \Appref{sec:more-pairing}.
% This pairing information changes the decision compared to \cpt{mean} and \cpt{median} in the middle figure.
% In \Figref{fig:geom_crit}~(b) and \Figref{fig:geom_crit}~(d), we see two instances where \cpt{BT} disagrees with \cpt{Mean} and \cpt{Median}.
% This graphical criterion captures the intuition that \cpt{BT} predicts as best the system that wins on most of the test instances.
% \todo{SE: In NLP (or ML in general), we could think of the case where the test data contains hard and easy cases. Let's say we have a case where one system is consistently better on the easy cases, but considerably worse on the hard cases. My feeling is that BT gets this wrong when the easy cases are in the majority}

\subsection{Connection with statistical testing}
\label{ssec:stat_test}
The above discussion about the differences between \cpt{Mean}, \cpt{Median}, and \cpt{BT} has interesting parallels with statistical testing.

When comparing the means of two systems over the same test set, the recommended statistical test is the \emph{paired} $t$-test \cite{fisher-1935}.
% When comparing the medians of two systems, the associated statistical test would be Mood's median test. But the data is paired, so instead, the recommended test should be the 
% and Mood's median test ignores the pairing \cite{wilcoxon_weak}.
% The paired alternative to the Mood's median test is the
When comparing medians instead of means, the appropriate test is the
sign test, which measures whether the median of the difference is significantly differerent from zero. Interestingly, the statistic of the sign test is precisely the one in the condition for \cpt{BT} to be consistent (see \Propref{prop:main_prop}). Wilcoxon's signed-rank test \cite{wilcoxon1945individual} is often used as an alternative to the sign test because it has more statistical power (at the cost of making more assumptions). However, \newcite{wilcoxon_weak} showed that Wilcoxon's signed-rank test does not always properly account for the pairing of data, unlike the sign test.

When performing statistical testing, it seems obvious that we should use the paired version of tests when the data is naturally paired \cite{rankel-etal-2011-ranking}. 
Even works discussing statistical testing in NLP recommend Wilcoxon's signed-rank test~\cite{Graham:2015,Owczarzak:2012, dror-etal-2018-hitchhikers}. Yet, to obtain aggregated scores for systems, the community still mostly uses aggregation mechanisms that ignore the pairing, such as \cpt{mean}. \cpt{MEDIAN} is the outlier\hyp resistant version of \cpt{mean}, and \cpt{BT} is the paired variant of \cpt{Median}.
Whenever one recommends a paired test of medians, such as the sign test or Wilcoxon's signed-rank test, to obtain $p$-values, one should use \cpt{BT} to compare system scores.

% \todo{Wei: what's the strength of BT over paired t-test and Mood's median test? MP: BT is not a statistical test so it is not comparable, the test related to BT is the sign test, the advantage of the sign test compared to Mood is that it is paired and compared to the paired t-test is that it is based on median, it requires less assumptions to be valid.}
% But we argued here that we could also consider paired aggregation mechanisms like \cpt{BT}. 
\section{Simulations with synthetic data}
\label{sec:simulations}
Next, we perform simulations to extend the analysis of the previous section to (i) $N>2$ systems, (ii) finitely many test samples, (iii) a practical implementation of \cpt{BT} (for $N>2$ systems, \cpt{BT} is an iterative optimization algorithm, as discussed in \Appref{app:bt_details}).

We synthesize evaluation scores with various properties starting with systems of predefined implicit strengths $\lambda_{i}$.
To create situations where the pairing of evaluation scores matters, we introduce multiple test instance types. For each type, systems perform differently but still have the same relative strength ($\mathbb{P}(A>B)$), differing only by an added offset. For example, the evaluation scores obtained by $A$ and $B$ could be sampled from $\mathcal{N}(\lambda_A, \sigma)$ and $\mathcal{N}(\lambda_B, \sigma)$ for one test instance type, and by $\mathcal{N}(\lambda_A + \epsilon, \sigma)$ and $\mathcal{N}(\lambda_B + \epsilon, \sigma)$ for another type, with $\epsilon$ being the offset.
% \todo{Wei: Maybe illustrate one example for test type.} 
% When several test types are considered we also include an \emph{outlier test type} where the relative strengths (given by $\lambda_i$) are not respected, by randomly selecting new $\lambda$'s just for this special test type.
We sample evaluation setups by varying the following properties: the number of systems, the number of test instances, the percentage of outliers, the numbers of test instance types, and the level of noise. This results in 2,880 simulated evaluation setups.
In \Appref{sec:details-simulations}, we give the detailed algorithm and parameters used to generate the data.
% When the number of test instance types is greater than one, it creates Bivariate Exchangeability Issues (BEI) as illustrated in \Figref{fig:geom_crit}.

% For more details about the exact sampling algorithm used to generate this data, see \Appref{sec:details-simulations}.

\begin{figure*}[t]
    \centering
        \includegraphics[width=0.96\linewidth]{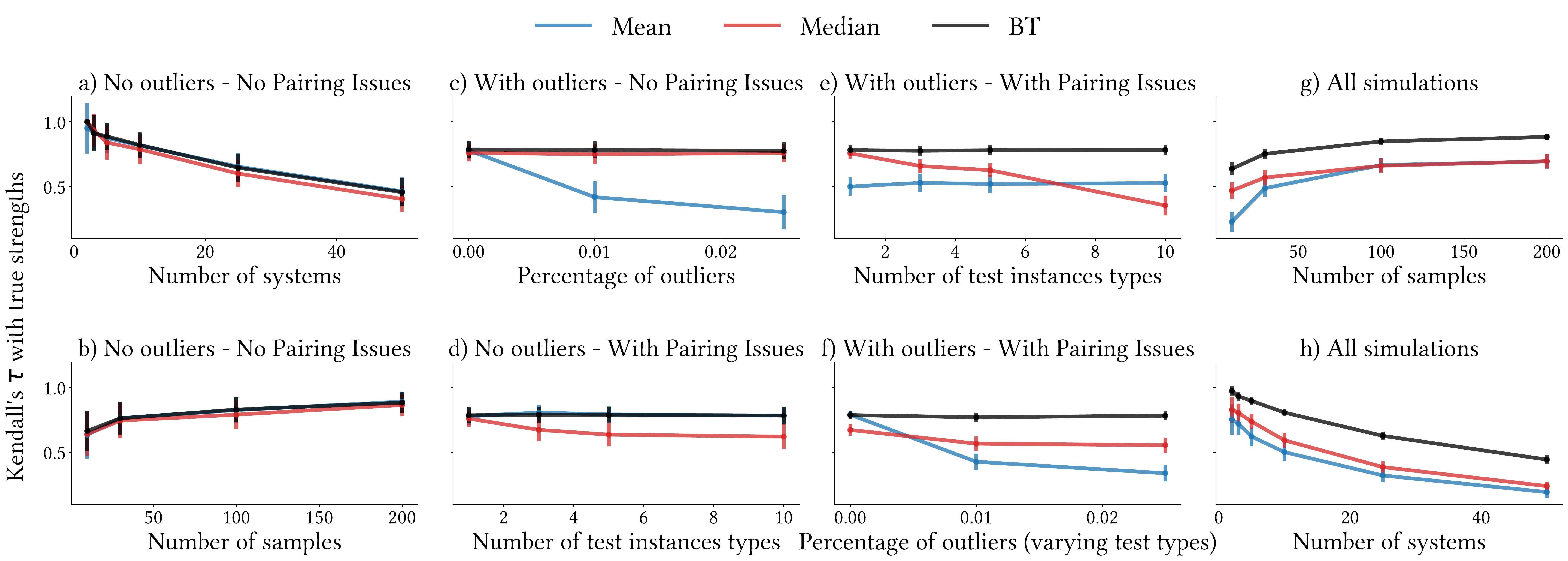}
        \caption{The y-axis is the Kendall's $\tau$ correlation between latent scores $\lambda_i$ of systems and the scores obtained after aggregating simulated evaluation scores with \cpt{mean}, \cpt{median}, or \cpt{BT}. 
        \Figref{fig:simulations}(a) and \Figref{fig:simulations}(b) corresponds to the intuitive case where no problem occurs (no outliers, no pairing issues).
        % depicted as a function of the number of systems and the number of test samples.
        \Figref{fig:simulations}(c) adds outlier problems only, and  \Figref{fig:simulations}(d) adds pairing issues only by increasing the number of types of test instances.
        \Figref{fig:simulations}(e) and (f) show the combined effect of outliers and pairing issues. Finally, \Figref{fig:simulations}(g) and \Figref{fig:simulations}(h) collect all the simulations. The error bars represent 95\% confidence intervals obtained with bootstrap resampling.
        }
        % and show the results as functions of the the number of systems and number of test samples.}
        \label{fig:simulations}
\end{figure*}

In \Figref{fig:simulations}, we report Kendall's $\tau$ between the latent scores $\lambda_{i}$ and the aggregated scores estimated by \cpt{mean}, \cpt{median}, and \cpt{BT}.
When the evaluation setup does not present any difficulty (\Figref{fig:simulations}(a,~b)), all aggregation mechanisms work equally well (within each other's $95\%$ error bounds), improving with more samples (\Figref{fig:simulations}(b)) and deteriorating with more systems (\Figref{fig:simulations}(a)). 
Unsurprisingly, \cpt{MEan} fails in the presence of outliers, whereas \cpt{median} and \cpt{BT} are unaffected (\Figref{fig:simulations}(c,~e,~f)).
When several types of test instances are considered, \cpt{median} begins to fail (\Figref{fig:simulations}(d)), which is made worse when outliers are also present (\Figref{fig:simulations}(f)).
Overall, \cpt{BT} is more robust and does not fail when the pairing matters \Figref{fig:simulations}(g,~h).
\section{Analysis of empirical data}
\label{sec:experiments}

In this section, we perform large-scale experiments using real evaluation scores from four NLG tasks.
For summarization, we use the TAC-08, TAC-09, TAC-11 and CNN/DM \cite{NIPS2015_afdec700} datasets.
For machine translation, we use the shared tasks of WMT-17 \cite{bojar-etal-2017-results}, WMT-18 \cite{ma-etal-2018-results}, and WMT-19 \cite{ma-etal-2019-results}. For image captioning, we use the MSCOCO \cite{Tsung-Yi-2014-coco} dataset, and for dialogue, we use the PersonaChat and TopicalChat \cite{mehri-eskenazi-2020-usr} datasets. 
The evaluation scores are obtained with a total of 18 different evaluation metrics:
BLEU-[1,2,3,4] \cite{Papineni:2002}, ROUGE-[1,2,L] \cite{Lin:2004}, ROUGE-WE-[1,2] \cite{ng-abrecht-2015-better}, JS-[1,2] \cite{lin-etal-2006-information}, S3-[pyr, resp] \cite{Peyrard:2017}, CIDEr \cite{VedantamZP15}, Chrfpp \cite{Popovic17}, METEOR \cite{Lavie:2007}, MoverScore \cite{zhao-etal-2019-moverscore}, and BERTScore \cite{bert-score}. Some metrics are only available for some task; e.g., CIDEr, METEOR are only available for the image captioning task.
We provide details about datasets, metrics, and their statistics in \Appref{sec:data_details}.

Overall, across datasets and metrics we have 296 evaluation setups, 73,471 pairs of systems, and 91,197 test instances. We also experiment with sub-sampling different sizes of test sets (see \Appref{sec:data_details}) to simulate varying train/dev/test splits or cross-validation.

\subsection{Comparison of \cpt{BT}, \cpt{mean}, and \cpt{median}}
In \Tabref{tab:global_results}, we report the disagreement between aggregation mechanisms over all the data with three measures: the percentage of pairs ranked in a different order (rescaled version of Kendall's $\tau$), the percentage of setups where the state-of-the-art (SotA) systems are different, and the percentage of setups where the top 3 systems are different (compared as sets).
A significant fraction of pairs of systems (about $10 \%$) are ranked differently by different mechanisms. More importantly, top systems are often different (in about $40\%$ of setups for top 1 and $50\%$ for top 3). We can conclude that the choice of aggregation mechanism has a real impact on evaluation outcome.
The observed disagreement between the three aggregation metrics implies that we are not in the case depicted by \Figref{fig:simulations}(a) and \Figref{fig:simulations}(b), i.e., the pairing matters and there are outliers in real data.
% As argued above, when the aggregation mechanisms disagree, \cpt{BT} is generally more trustworthy because it is robust to outliers and utilizes the natural pairing of the data.
In the next paragraphs, we break down the disagreement per evaluation metric, task, and test set size. Detailed results are provided in \Appref{app:disagree_breakdown}.
% We also show concrete example of disagreement and associated pair-plots in \Appref{app:example_pairs}.
% \se{Our main findings are:}

\begin{table}
\centering
\resizebox{0.80\columnwidth}{!}{
\begin{tabular}{ l | c c c }
\toprule
& Disagree & $\neq$ SotA & $\neq$ Top-3 \\
\midrule
\cpt{Mean} vs.\cpt{Median}   & $4\%$ & $18\%$ & $30\%$ \\
\cpt{Mean} vs. \cpt{BT}       & $9\%$ & $40\%$ & $49\%$ \\
\cpt{Median} vs. \cpt{BT}     & $9\%$ & $41\%$ & $55\%$ \\
\bottomrule
\end{tabular}
}
\caption{Disagreement between aggregation mechanisms. The first column shows the percentage of system pairs ordered differently by two aggregation mechanisms. The second column shows the percentage of setups where two aggregation mechanisms find different SotA, and the third column shows the percentage of setups where the top-3 systems are different (compared as sets).}
\label{tab:global_results}
\end{table}     

% \begin{itemize}
%     \item We find that a significant fraction of pairs of systems are ranked differently by different mechanisms
%     \item more importantly, top systems are often different. This means that choosing an aggregation mechanism has an important impact on the directions taken by the community. %research field
%     \item The disagreement between the different metrics indicate that synthetic examples shown in previous sections actually happen in real data.
%     \item Because aggregation mechanisms disagree, we are not in the situation depicted by \Figref{fig:simulations}(a) and \Figref{fig:simulations}(b).
%     \item As argued above, when there is disagreement BT is more trustworthy because it is robust to both outliers and pairing problems.
% \end{itemize}

\begin{figure*}[t]
    \centering
    \includegraphics[width=0.79\linewidth]{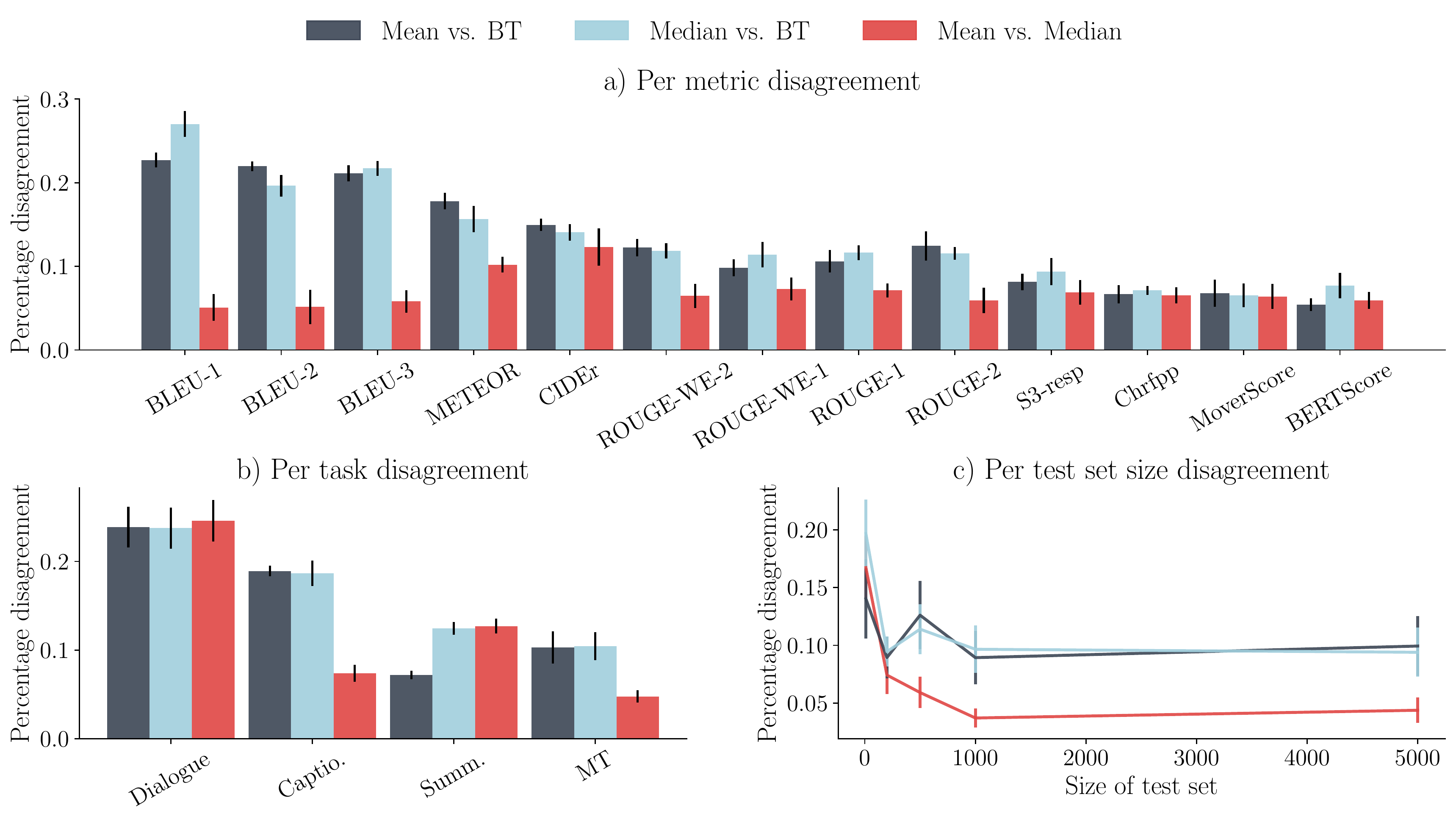}
    \caption{This figure measures the percentage of disagreement between each pair of aggregation mechanisms across different dimensions with real evaluation setups. \Figref{fig:impact-all}(a) shows the disagreement per evaluation metric averaged over tasks and uniformly subsampled test set sizes, \Figref{fig:impact-all}(b) shows the disagreement per task averaged over evaluation metrics and uniformly subsampled test set sizes, and \Figref{fig:impact-all}(c) shows the disagreement across test set sizes averaged over tasks and evaluation metrics.}
    \label{fig:impact-all}
\end{figure*}

\paragraph{Which metrics are impacted most?}
% Now, we ask which metric suffers most from the discrepancies between the aggregation mechanisms.
We report in \Figref{fig:impact-all}(a) the percentage of disagreement between aggregation mechanisms per metric averaged over datasets, when subsampling test sets of different sizes uniformly (see \Appref{sec:data_details} for details).
While most metrics are available for all four tasks, METEOR and CIDEr are only available for the captioning task. Therefore, the observed disagreements for these metrics may be a feature of the task instead of the metrics.
% \todo{SE: are these two different cases - sample uniformly and subsampling? What does it mean to sample the dataset uniformly?}
% Different metrics are affected differently because they may induce different properties on the score distributions.
Interestingly, recent metrics such as BERTScore and MOVERScore seem less affected. On the other hand, BLEU variants are the most impacted, particularly when comparing \cpt{MEAN} or \cpt{MEdian} against \cpt{BT}. The disagreement between \cpt{Mean} and \cpt{MEdian} is stable across metrics. In general, \cpt{MEAN} and \cpt{Median} are more in agreement with one another than they are with \cpt{BT}, which indicates that pairing issues have a stronger effect than outliers.

% In comparison to ROUGE or BLEU, their outputs are really continuous creating fewer outliers

% \begin{itemize}
%     \item We observe that different metrics are affected differently because each metric induces different properties on the distribution of scores.
%     \item Interestingly, recent metrics like BERTScore and MoverScore seem less affected. This is because they are really continuous and do not match all poor systems to $0$, creating fewer outliers.\todo{SE: The last sentence is unclear to me} 
%     \item BLEU variants are the most affected with up to $30\%$ of the pairs ordered differently between the median and BT.
%     \item mean and median are more in agreement together than they are with BT. It indicates that the pairing issues are more important than the presence of outliers.
% \end{itemize}

% We report the percentage of disagreement between each pairs of aggregation: mean vs. median, mean vs. BT, and median vs. BT.

% (We also have other measures of disagreement, e.g., kendall's $tau$ which is equivalent to this one, or the percentage of time the SotA is different, or whether top n systems are the same, ...)

\paragraph{Which tasks are impacted most?}
% We now ask which tasks suffers most of these discrepancies.

\Figref{fig:impact-all}(b) summarizes an analysis as above, but across tasks instead of metrics. Again, to control for the fact that some tasks may have larger datasets, we subsample uniformly from various test set sizes. The results are averaged over evaluation metrics.
% We also select randomly  from the set of metrics.
Machine translation and summarization suffer the least while dialogue and image captioning display larger disagreement between aggregation mechanisms. This suggests important future research directions to improve the evaluation setups in these tasks. 
% Again, we see that \cpt{mean} and \cpt{median} agree more together than they agree with BT.

% \begin{itemize}
%     \item MT suffers the least, yet previous works that used ranking metrics were most often applied to MT and not other tasks.\todo{SE: the "yet" part is unclear to me. You are not going to say that the MT folks have used our pairwise aggregation techniques already, are you?} It seems that other tasks would greatly benefit from\todo{SE: greatly benefit from?} 
%     \item Image captioning is particularly suffering, which suggests important future research in the proper evaluation of image captioning systems.\todo{SE: actually, dialogue suffers the most}
%     \item We again see that that mean and median agree more together than they agree with BT.
% \end{itemize}

\paragraph{Importance of dataset size.}
% Finally, we investigate the importance of test set size, as we already saw in simulation \Figref{fig:simulations} that the number of test instances plays in an important role.

In \Figref{fig:impact-all}(c), we report disagreement across test set sizes, while averaging over datasets and evaluation metrics.
It is reassuring to observe that with larger test sets, the different mechanisms tend to agree more, such that it matters less which one is actually chosen. However, for \cpt{Mean} vs.\ \cpt{BT} and \cpt{median} vs.\ \cpt{BT}, the disagreement does not continue to decrease below $10\%$ with more test instances.
% it never falls below $10\%$ disagreement. 
For \cpt{Mean} and \cpt{BT} the disagreement is lower but exhibits the same behavior, never falling below a certain threshold. %$5\%$. 
% It indicates fundamental differences in the data distributions and pairing patterns, as discussed in \Secref{sec:framework}.  

\paragraph{Different perspectives on uncertainty.}
In standard evaluation setups, not only system scores are reported but also whether the differences are statistically significant \cite{dror-etal-2018-hitchhikers}. Therefore, we ask how often differences that are statistically significant for one test are also statistically significant for another. The details of this experiments are presented in \Appref{app:uncertainty} and show, perhaps unsurprisingly, different behavior for different tests. In particular, the paired $t$-test is the one that most often finds differences to be significant (for $41\%$ of pairs); Mood's test, an unpaired test to compare medians, finds significance for only $21\%$ of pairs; and the sign test and Wilcoxon's sign-rank test (related to \cpt{BT}) are in between (for $35\%$ and $40\%$ of the pairs, respectively).

\paragraph{Sources of disagreement.}
Based on the analysis of \Secref{sec:theory}, we know that the difference between \cpt{Mean} and \cpt{Median} is due to the presence of statistical outliers, while the difference between \cpt{Median} and \cpt{BT} is due to the presence of different test instance types (\Figref{fig:simulations}). With real NLP datasets, in \Figref{fig:impact-all}, we observe some discrepancy between \cpt{Mean} and \cpt{Median}, indicating the presence of outliers. There is even more disagreement between \cpt{Median} and \cpt{BT}, indicating the presence of different types of test instances, as illustrated in \Figref{fig:simulations}.

\section{Related work}

% \xhdr{Meta-Evaluation}
Several studies have made a critical assessment of the standard evaluation methodologies. For example, \newcite{freitag-etal-2020-bleu} demonstrate the advantages of carefully choosing which references to use for NLG evaluation. 
% \newcite{peyrard-2019-studying} showed that the choice of evaluation metric matters.
% Also for MT, 
\newcite{mathur-etal-2020-tangled} 
% reports that current model selection methods are highly sensitive to the translations used for assessment. They 
show that outliers matter in practice.
Recently, \newcite{graham-etal-2020-statistical} draws attention on test set size.
% and statistical power in MT. 
% Also, \cite{schuff-etal-2020-f1} discusses shortcomings of F1.
Several works have emphasized the importance of careful statistical testing \cite{rankel-etal-2011-ranking, Owczarzak:2012,Graham:2015,dror-etal-2018-hitchhikers}. They recommend \textit{paired} statistical tests.
% like the paired t-test or the Wilcoxon sign-rank test.
% More related to our discussion, \newcite{rankel-etal-2011-ranking} discussed the importance of statistical testing and mentioned ranking-based metrics.
% In the context of classification,  
% \cite{schuff-etal-2020-f1} evaluation settings have shortcomings regarding the coupling of answer and explanation which might cause serious issues
% \newcite{Owczarzak:2012}, \newcite{Graham:2015}, and \newcite{dror-etal-2018-hitchhikers} re-evaluates statistical testing in NLP and call for more careful testing. 
% Also for the machine translation task, \newcite{mathur-etal-2020-tangled} reports that current model selection methods are highly sensitive to the translations used for assessment. They show that outliers matter in practice.
Finally, \citet{novikova-etal-2018-rankme} report that ``relative rankings yield more discriminative results than absolute assessments''%
% when evaluating natural language generation
, which further motivates aggregation mechanisms like \cpt{BT}.
% These works provide additional motivation for the use of robust aggregation mechanisms like \cpt{BT}.

% \cite{Lin:2003:ncc} gave two criteria that evaluation methodologies should meet: a) correlate highly, positively, and consistently with human assessments.
% b) the statistical significance of automatic evaluations should be a
% good predictor of the statistical significance of human assessments

% ,\todo{SE: reference translations? Could this be 
% addressed by avoiding them?} particularly the presence of outliers, which often leads to falsely confident conclusions about a metric’s efficacy. Show that outliers matter in practice for the evaluation of evaluation metrics in MT and discuss robust alternatives like Wilcoxon signed-rank test.

\xhdr{Aggregations}
Pairwise comparison mechanisms date back to \newcite{Thurstone}. Subsequently, the Bradley-Terry (BT) model has become a standard pairwise comparison model \cite{bt_one}. 
% \newcite{cattelan2012} studied extension in the case of dependent data.
In NLP, BT-inspired mechanisms have sometimes been used to aggregate human assessments.
% In NLP, these have rarely been used to replace average, other than for aggregating human assessments. 
For instance, \newcite{deriu-etal-2020-spot} ranked chatbots regarding their ability to mimic conversational behavior of humans. 
% \newcite{lombardo-etal-2020-top} used ranking correlation coefficients to evaluate semantic models
% via the aforementioned dataset by taking into account the greater significance of top ranks
% Some variants of BT have also been proposed and discussed. 
Item response theory (IRT) has a similar formulation as BT, but also estimates the difficulty of each test instances using a latent\hyp variable Bayesian model \cite{dras-2015-squibs}. IRT has been applied to perform dataset filtering \cite{lalor-etal-2016-building,lalor-etal-2019-learning}, evaluate chatbots from human assessments \cite{sedoc-ungar-2020-item}, and aggregate human assessments in machine translation \cite{dras-2015-squibs}.
Elo \cite{elo1978rating} and TrueSkill \cite{true_skill} are famous extensions of the BT model commonly used to rate players in the context of gaming or sports events.
Elo views player strengths as normally distributed random variables. TrueSkill is a Bayesian variant of Elo. 
Since 2015, the Workshop on Machine Translation (WMT) has been using TrueSkill to rank models based on human assessments following the methodology of \newcite{sakaguchi-etal-2014-efficient}.
We provide a detailed presentation and comparison of BT, Elo, and TrueSkill in \Appref{app:variants_bt}, and make both Elo and TrueSkill available as alternatives to \cpt{BT} in the released tool. The arguments in favor of \cpt{BT} made in this work transfer to its variants, including IRT, Elo, and TrueSkill, and the conclusions drawn from the experiments of 
% While \cpt{BT} is simpler and has no parameters, Elo and TrueSkill can be updated online, as new instances arrive.
\Secref{sec:experiments} still hold when replacing \cpt{BT} by Elo or TrueSkill (\Appref{app:variants_bt}).
Our work extends previous works that has considered \cpt{BT} variants by analyzing the potential causes for disagreement with \cpt{MEAN} and \cpt{median} and by measuring the disagreement in real NLP evaluation setups.

\section{Discussion}
\label{sec:discussion}

We briefly discuss some possible questions raised by the use of \cpt{BT}-like metrics, with more details provided in \Appref{app:bt_details}, \ref{app:arrow_theorem}, \ref{app:variants_bt}, and \ref{app:repeating_experiments}.

\xhdr{Extension to other evaluation setups}
The experiments of \Secref{sec:experiments} focus on reference-based NLG evaluation metrics. However, the arguments laid out throughout the paper apply beyond this setup. Any comparison of systems based on score aggregation is susceptible to suffer from outliers and complex pairing structures (e.g., \Figref{fig:geom_crit}). Future work should replicate our experimental setup for reference-free NLG \cite{zhao-etal-2020-limitations}, classification, or regression tasks.

% In our experiments with real data, we focused on NLG evaluation metrics and tasks. However, the arguments laid out in favor of paired evaluation throughout the paper are general and apply beyond the scope of NLG. Any comparison of systems based on an aggregation over is susceptible to suffer from outliers and a complex pairing structure (e.g., \Figref{fig:geom_crit}). In particular, reference-free NLG metrics , classification, and regression tasks

\xhdr{Type imbalance}
Imagine a test set with a majority of easy instances and few hard ones. A system $A$ could perform slightly worse than $B$ on easy instances but much better on hard ones and will be declared worse by \cpt{BT}.
If one views this decision as problematic then one should probably acknowledge that the test set is not representative of what should be measured. If hard instances matter more there should be a majority of them in the test set. Hoping that \cpt{mean} will be swayed to output the \emph{intuitive} ordering of systems from a minority of test instances is a peculiar expectation to have about the evaluation setup.
To diagnose such pathological cases, our tool, \textit{Pairformance,} offers the possibility to view pairwise plots (as in \Figref{fig:geom_crit}) and histograms of score differences.
More generally, better aggregation mechanisms such as \cpt{BT} do not solve all potential problems of evaluation methodologies. Other aspects (such as choosing evaluation metrics or meaningful, representative, and large test sets) are all independent of the choice of aggregation mechanism, but also critical to the quality of the evaluation.

\xhdr{Transitivity}
\cpt{BT} is not computed independently for each system, and it can happen that adding or removing a baseline impacts the scores of other systems. We explain this phenomenon in \Appref{app:arrow_theorem} and show that it is rarely a problem in real data. More generally, we discuss the connection with Arrow's impossibility theorem in the context of the aggregation of social preferences \cite{arrow1950difficulty}. The \textit{Pairformance} tool gets around this difficulty by offering the possibility of analyzing each pair of systems independently.

\xhdr{Relaxing assumptions}
\cpt{BT} assumes that the relative strengths of systems remain constant across test instances. This might not always be true, especially when some systems are crafted for some specific kind of instances but perform badly on others. In such cases, \cpt{BT} still produces meaningful and easily interpretable results but fails to capture the latent structure of system strengths. Several refinements of \cpt{BT} are possible; \eg, item response theory extends \cpt{BT} by modeling instance difficulty, and Elo and TrueSkill allow system strengths to be stochastic and vary across instances. These refinements come at the cost of introducing new parameters, and it remains unclear how to choose these parameters in practice. Future work should investigate systematic ways to choose these parameters.

\xhdr{Tool description}
% \label{app:tool}
% To further motivate the community to use ranking-based metrics to evaluate systems, we release a user-friendly tool.
We release \textit{Pairformance,} a tool for performing full diagnostic analyses based on an evaluation dataframe made of the evaluation scores of systems and baselines.
% with the chosen evaluation metrics on some test set of interest.
\begin{figure}
    \centering
    \includegraphics[width=0.65\linewidth]{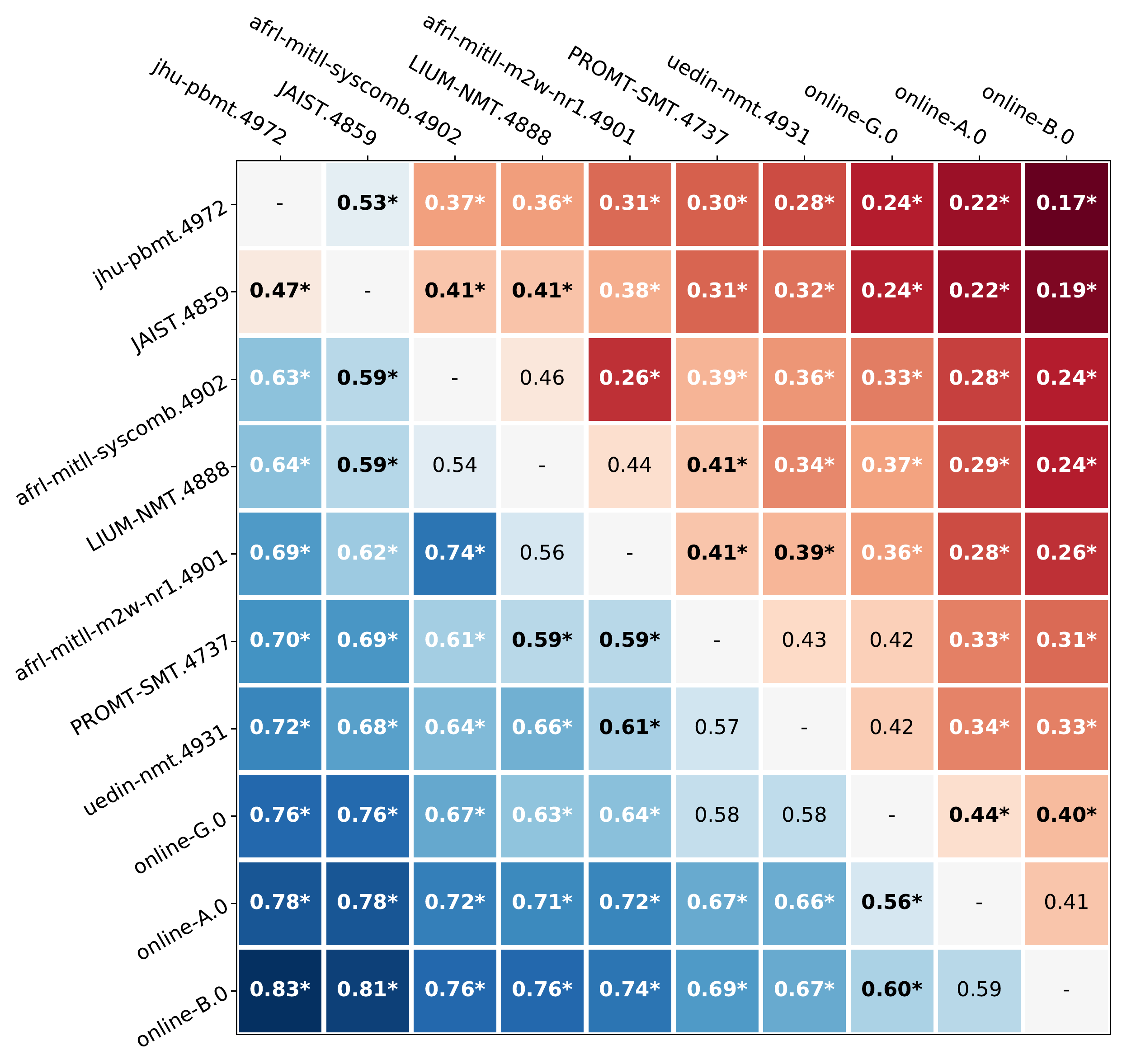}
    \caption{Pairwise system comparison with \cpt{BT} for machine translation with ROUGE-1, as output by the \textit{Pairformance} tool released as part of this work.}
    \label{fig:pairwise-system-comparison}
\end{figure}
It can perform the analysis based on \cpt{Mean}, \cpt{Median}, \cpt{BT}, Elo, and TrueSkill. 
For each aggregation technique, it outputs a full pairwise analysis of all pairs of systems. 
For \cpt{Mean} and \cpt{MEdian} it compares score differences for pairs of systems. For \cpt{BT}, Elo, and TrueSkill, it estimates the probability that one system is better than another. All analysis is accompanied by appropritate statistical testing. See \Figref{fig:pairwise-system-comparison} for an example based on the \cpt{BT} mechanism.
Furthermore, the tool can plot the histogram of paired differences $X_A^{(l)} - X_{B}^{(l)}$, allowing for the direct identification of pathological patterns such as those discussed above.

\section{Conclusion}
We performed a critical assessment of the standard NLP evaluation methodology based on averaged scores, which ignores the natural instance\hyp level pairing of evaluation scores when comparing systems. 
We showed the importance of the pairing and demonstrated the advantages of paired mechanisms such as Bradley--Terry (\cpt{BT}) over more standard aggregation schemes such as the mean or median. The choice of aggregation mechanism matters in real evaluation setups, and we therefore recommend \cpt{BT} as a robust aggregation mechanism. To facilitate adoption, we release \textit{Pairformance,} a new tool to perform full analyses of system scores using \cpt{BT} and two of its variants, Elo and TrueSkill.

\section*{Acknowledgments}
We thank the anonymous reviewers for their insightful comments and suggestions, which greatly improved the final version of the paper. 
With support from Swiss National Science Foundation (grant 200021\_185043), European Union (TAILOR, grant 952215), and gifts from Google, Facebook, Microsoft.

\bibliography{acl2020,wei}
\bibliographystyle{acl_natbib}

\clearpage
\appendix
\section{Reproducibility}

In this section, we give additional details to ensure the reproducibility of our experiments. Furthermore, the code and data to reproduce each figure and table of the main paper is available at: \url{https://github.com/epfl-dlab/BT-eval}.

\subsection{Pairing examples}
\label{sec:paring-examples}

It is straightforward to generate examples where the marginal distribution of the evaluation scores of two systems remain unchanged even when the pairing varies.

To do so, one can define $k$ types of test instances.
For each type $t_i$, each system has a probability distribution of scores for this type: $\mathcal{N}(\lambda_{t_i, S}, 1)$.
So for instances of type $t_i$, the system $S$ has score $\lambda_{t_i, S}$ in expectation with a variance of $\sigma^2 = 1$. 
Similarly, another system $B$ can have different $\lambda_{t_i, B}$ parameters. An example is given in \Tabref{tab:example-test-types}. 

\begin{table}
        \small
        \centering
        \resizebox{0.85\columnwidth}{!}{
        \begin{tabular}{l|ccccc}
        \toprule
         & type 1 & type 2 & type 3 & type 4 & type 5 \\
        \midrule
        $S$     & 23 & 50 & 40 & 70 & 60 \\
        $B$     & 28 & 45 & 30 & 65 & 50 \\
        \bottomrule
        
        \end{tabular}
        }
        \caption{Example of two systems $S$ and $B$ with their strengths $\lambda_{t_i, S}$ and $\lambda_{t_i, B}, i \in [1, 5]$ associated to each type of test instances. types.}\label{tab:example-test-types}
\end{table}

Now, observe that permuting the columns of $S$ without changing the row $B$ leaves the marginal distribution of $S$ and $B$ unchanged but changes the pairing. Then, one can simply iterate over all permutations of the row $S$ to obtain many different pairings with fixed marginal distributions. 
% For this particular example, we report 25 pairings (out of the $5! = 120$ possible) in \Figref{fig:example-bei}. In all of these plots, the marginal distribution of the scores of $S$ and $B$ remain unchanged and, thus, their mean and median (and any other statistic of their distribution scores only) remain unchanged.

\subsection{Simulation}
\label{sec:details-simulations}
% The strategy to generate the pathological synthetic data examples depicted in \Figref{fig:geom_crit} is detailed in the next section \Appref{sec:more-pairing}. 
We discuss the synthetic data and experiments depicted in \Figref{fig:simulations}.

To introduce pairing issues, we create a variable number of test instance types: $N_{types}$.
For each test type, each system has a different distribution of scores.
On test type $t_i$, the system $s_j$ has a normal distribution of scores: $\mathcal{N}(\lambda_{i,j}, \sigma^2)$, where we fix $\sigma^2 =1$ throughout our experiments. For each system, the $\lambda_{i,j}$ are sampled uniformly from $[0, 1]$. Depending on the values of $\lambda_{i,j}$, the score distribution of system $s_j$ can become multimodal.
When, there is only one test type, the score of each system $s_j$ is a normal $\mathcal{N}(\lambda_j, \sigma^2)$. In that case, the pairing can be ignored and \cpt{MEAN} and \cpt{MEDIAN} are expected to work well.

For outliers, we define $f$ as the fraction of test instances on which systems' scores are not drawn from their distribution scores. For such instances, we first draw the scores for each systems according to their distribution and then perform a random permutation, so that each system receives a score that is not sampled from its score distribution.

Then, we vary the number of systems present in the evaluation $N_{sys}$ and the number of test instances $M$.
Each choice of $N_{types}, f, N_{sys}, $ and $M$ gives a dataframe corresponding to an evaluation setup on which we can compare \cpt{MEAN}, \cpt{MEDIAN}, and \cpt{BT} against the \emph{true} latent strengths of systems $\lambda_{i, j}$.
The evaluation and the y-axis in \Figref{fig:simulations} is then the Kendall's $\tau$ between the ordering resulting from \cpt{MEAN}, \cpt{MEDIAN}, or \cpt{BT} against the ordering resulting from the $\lambda_{i, j}$. 

We consider the following variations for the parameters of the experiments:
\begin{itemize}
\denselist
    \item $N_{types} \in \{1, 3, 5, 10\}$,
    \item $f \in \{0., 0.01, 0.025\}$,
    \item $N_{sys} \in \{2, 3, 5, 10, 25, 50\}$,
    \item $M \in \{10, 30, 100, 200\}$.
\end{itemize}
In total, we have: $4 \cdot 3 \cdot 6 \cdot 4 = 288$ parameter choices. 
For each we sample $10$ datasets resulting in $2,880$ synthetic evaluation setups.

\subsection{Real data}
\label{sec:data_details}
% We collected datasets from $4$ important NLP tasks: summarization (4 datasets), machine translation (3 datasets), image captioning (1 dataset), and dialogue (2 datasets).
Each of the dataset we use contains the evaluation results of a varying number of systems for a varying number of evaluation metrics:

% \begin{itemize}
% \denselist
\textbf{Summarization}:
    CNN/DM \cite{NIPS2015_afdec700}: 11,432 test instances, 12 summarization systems, and 13 evaluation metrics.
     TAC-08: 48 test instances, 58 summarization systems, and 13 evaluation metrics.
     TAC-09: 44 test instances, 55 summarization systems, and 13 evaluation metrics.
 TAC-11: 44 test instances, 50 summarization systems, and 13 evaluation metrics.
 \textbf{Captioning}:
    MSCOCO \cite{Tsung-Yi-2014-coco}: 40,504 test instances, 12 systems, and 7 evaluation metrics.
\textbf{Dialogue}: Topical-Chat \cite{mehri-eskenazi-2020-usr}: 60 test instances, 5 systems, and 13 evaluation metrics.
    Persona-Chat \cite{mehri-eskenazi-2020-usr}: 60 test instances, 4 systems, and 13 evaluation metrics.
    \textbf{MT}: 
    WMT-17 \cite{bojar-etal-2017-results}: evaluated with 11 evaluation metrics, we have the following pairs: \emph{lv-en} (2,001 instances, 9 systems), \emph{de-en} (3,004 instances, 11 systems), \emph{ru-en} (3,001 instances, 9 systems), \emph{tr-en} (3,007 instances, 10 systems), and \emph{zh-en} (2,001 instances, 16 systems).
    WMT-18 \cite{ma-etal-2018-results}: evaluated with 13 evaluation metrics we have the following pairs: \emph{de-en} (2,998 instances, 16 systems), \emph{et-en} (2,000 instances, 14 systems), \emph{fi-en} (3,000 instances, 9 systems), \emph{ru-en} (3,000 instances, 8 systems), and \emph{zh-en} (3,981 instances, 14 systems).
    WMT-19 \cite{ma-etal-2019-results}: evaluated with 13 evaluation metrics we have the following pairs: \emph{de-en} (2,000 instances, 16 systems), \emph{fi-en} (1,996 instances, 12 systems), \emph{gu-en} (1,016 instances, 12 systems), \emph{kk-en} (1,000 instances, 11 systems), \emph{lt-en} (1,000 instances, 11 systems), \emph{ru-en} (2,000 instances, 14 systems), and \emph{zh-en} (2,000 instances, 15 systems).

The evaluation metrics considered are:
BLEU-[1,2,3,4] \cite{Papineni:2002}, ROUGE-[1,2,L] \cite{Lin:2004}, ROUGE-WE-[1,2] \cite{ng-abrecht-2015-better}, JS-[1,2] \cite{lin-etal-2006-information}, S3-[pyr, resp] \cite{Peyrard:2017}, CIDEr \cite{VedantamZP15}, Chrfpp \cite{Popovic17}, METEOR \cite{Lavie:2007}, MoverScore \cite{zhao-etal-2019-moverscore}, and BERTScore \cite{bert-score}. This is a total of 18 metrics. 
% Note that some metrics are only available for some task, e.g., CIDEr, METEOR are only available for the image captioning task.

% Each evaluation setup results in a dataframe. These dataframe are used in the analysis of \Secref{sec:experiments}. 
% We release the code to generate the preprocessed dataframe from existing sources used to generate figure and table of the paper at: \url{anonymized-link}.

\xhdr{Sub-sampling test set sizes}
In experiments reported by \Figref{fig:impact-all} the results are averaged after resampling test sets of different sizes.
The test set sizes used are: $[10, 50, 100, 500, 1000, 5000]$.
% In \Figref{fig:impact-all}~(a) and (b), this resampling controls for test set sizes to focus the difference on the metric and on the task by removing the dependency between task and dataset sizes. In the case of the dialogue task, where there is no dataset with $5000$ test instances, we upsample the existing dataset to simulate larger statistical power.
Results broken down per dataset and per metric that does not need resampling of test set sizes is proposed in \Appref{app:disagree_breakdown}.

\subsection{Implementations}
We implement \cpt{BT} with \url{scipy.org} and numpy.
% The implementation of \cpt{BT} is rather straightforward as it consists in iteratively minimize the log-likelihood:
For the statistical tests, we use the default implementation from \url{scipy.org}.
For Elo, we implement a wrapper around existing code: \url{https://github.com/ddm7018/Elo}.
Similarly, for TrueSkill, we implement a wrapper around existing code: \url{https://pypi.org/project/trueskill/}.

% Our implementation of \cpt{BT} and wrapper around Elo and TrueSkill are all integrated in the comparison tool we release. 
% The capabilities of this tool is described in \Appref{app:tool}.

\section{Proof of Proposition 1}
\label{sec:proof}
% First, we remind the statement of the proposition:
% If $\lambda_A > \lambda_B$:

% \begin{itemize}
%     \item \cpt{Mean} consistent $\iff  E_A -  E_B > 0$
%     \item \cpt{Median} consistent $\iff M_A - M_B > 0$.
%     \item \cpt{BT} consistent $\iff M_{A-B} > 0$,
% \end{itemize}
% where $E_S$, and $M_S$ are random variables obtained from taking the mean and median of the list of random variables from the evaluation scores of system $S$. $M_{A-B}$ is the median of the differences between the evaluation scores of $A$ and $B$.

\begin{proof}
We observe that the case of the \cpt{MEan} and the \cpt{Median} are direct by definition.
% , considering that, for \cpt{Mean}, the score of system $A$ is given by $E_{A}$ and \cpt{mean} declares $A$ stronger if it scores higher than $B$, i.e., $ E_A -  E_B > 0$. Replacing the average by the median gives the second point.

$M_{A-B}> 0$ is equivalent to saying that for more than $50\%$ of instances, $X_A^{(l)} > X_B^{(l)}$, i.e., $A$ is better than $B$ on more than $50\%$ of instances.
On the other hand, \cpt{BT} correctly gives $A$ better than $B$ $\iff \mathbb{P}(A > B) > \mathbb{P}(B > A) \iff \mathbb{P}(A > B) > \frac{1}{2}$, i.e., $A$ is better than $B$ on more than $50\%$ of instances.
So, \cpt{BT} is consistent $\iff$ $A$ is better than $B$ on more than $50\%$ of instances $\iff$ $M_{A-B}> 0$.
\end{proof}

\section{Disagreement breakdown}
\label{app:disagree_breakdown}
% In the main paper \Figref{fig:impact-all}, we report disagreement aggregated by metrics while uniformly sampling tasks and subsampling test set sizes, disagreement aggregated by tasks while uniformly sampling metrics and subsampling test set sizes, and disagreement per test set sizes while uniformly sampling metrics and tasks.
Compared to experiments in the main paper, we provide a more 
% However, not all tasks are evaluated by all metrics in the, so we provide a more 
detailed breakdown of the disagreement in \Tabref{tab:breakdown_disagreement}. 
% Furthermore, instead of just reporting the disagreement, we also report the percentage of time the SotA is different and the percentage of the top 3 systems are different.

% Finally, since it still does not cover the full cross-product between datasets and metrics (288 cases) we release, with the code to reproduce our results, a simple function computing the disagreement, percentage different SotA, and percentage of different top3 of \cpt{mean} vs. \cpt{BT}, \cpt{mean} vs. \cpt{median}, and \cpt{median} vs. \cpt{BT} for any dataset and metric given as input. 

\begin{table*}[ht]
\centering
\resizebox{0.89\textwidth}{!}{
\begin{tabular}{ l l | c c ccccccccccccccc}
\toprule
& & \multicolumn{3}{c}{BLEU} &  \multicolumn{3}{c}{ROUGE} & \multicolumn{3}{c}{ROUGE-WE} & \multicolumn{3}{c}{MoverScore} & \multicolumn{3}{c}{BERTScore} \\

& & Mean/BT & Med/BT & Mean/Med & Mean/BT & Med/BT & Mean/Med& Mean/BT & Med/BT & Mean/Med& Mean/BT & Med/BT & Mean/Med& Mean/BT & Med/BT & Mean/Med\\
\midrule
\multirow{3}{*}{TAC08} 
& Disagree. &  .09 & .13 & .15 &  .07 & .13 & .14 &  .12 & .06 & .13 &  .05 & .11 & .12 &  .05 & .11 & .12 \\
& $\neq$ SotA &  .43 & .73 & .47 &  .33 & .52 & .47 &  .58 & .20 & .47 &  .10 & .50 & .47 &  .13 & .17 & .27  \\	
& $\neq$ Top3 &  .73 & .77 & .77 &  .61 & .80 & .81 &  .87 & .65 & .80 &  .43 & .73 & .70 &  .60 & .93 & .87 \\
\midrule
\multirow{3}{*}{TAC09} 
& Disagree. &  .08 & .13 & .13 & .08 & .16 & .16 & .07 & .15 & .16 & .06 & .14 & .13 &  .06 & .12 & .12 \\
& $\neq$ SotA & .00 & .00 & .00 & .00 & .00 & .00 & .00 & .00 & .00 & .00 & .00 & .00 &  .00 & .00 & .00 \\
& $\neq$ Top3 & .70 & .70 & .70 & .63 & .87 & .82 & .48 & .73 & .75 & .33 & .70 & .70 &  .43 & .73 & .67 \\
\midrule
\multirow{3}{*}{TAC11} 
& Disagree. &  .07 & .12 & .12 & .06 & .13 & .12 & .05 & .13 & .12 & .04 & .11 & .10 &  .04 & .11 & .10 \\
& $\neq$ SotA & .37 & .67 & .50 & .42 & .64 & .61 & .33 & .67 & .65 & .40 & .63 & .63 &  .27 & .73 & .63 \\
& $\neq$ Top3 & .73 & .87 & .83 & .58 & .88 & .87 & .60 & .93 & .92 & .57 & .87 & .80 &  .43 & .87 & .83 \\
\midrule
\multirow{3}{*}{CNN/DM} 
& Disagree. &  .14 & .17 & .12 & .08 & .07 & .02 & .06 & .05 & .02 & .07 & .06 & .08 &  .08 & .08 & .04 \\
& $\neq$ SotA & .53 & .80 & .83 & .00 & .00 & .00 & .00 & .00 & .00 & .00 & .00 & .00 &  .00 & .00 & .00 \\
& $\neq$ Top3 & .97 & .97 & .90 & .73 & .49 & .24 & .90 & .42 & .48 & .00 & .00 & .00 &  .90 & .90 & .06 \\
\midrule
\multirow{3}{*}{WMT17} 
& Disagree. &  .07 & .08 & .05 & .07 & .07 & .04 & .07 & .08 & .04 & .03 & .04 & .03 &  .03 & .04 & .03 \\
& $\neq$ SotA & .17 & .19 & .14 & .28 & .42 & .23 & .35 & .40 & .19 & .22 & .15 & .24 &  .15 & .22 & .24 \\
& $\neq$ Top3 & .43 & .57 & .40 & .56 & .63 & .29 & .57 & .67 & .40 & .26 & .37 & .37 &  .23 & .27 & .33 \\
\midrule
\multirow{3}{*}{WMT18} 
& Disagree. &  .09 & .09 & .03 & .11 & .11 & .04 & .12 & .12 & .04 & .06 & .06 & .04 &  .06 & .06 & .03 \\
& $\neq$ SotA & .67 & .63 & .24 & .55 & .65 & .26 & .61 & .67 & .66 & .47 & .49 & .18 &  .43 & .47 & .31 \\
& $\neq$ Top3 & .77 & .74 & .25 & .56 & .69 & .39 & .66 & 77 & .40 & .57 & .58 & .33 &  .57 & .58 & .19 \\
\midrule
\multirow{3}{*}{WMT19} 
& Disagree. &  .07 & .08 & .04 & .10 & .11 & .04 & .11 & .11 & .05 & .05 & .04 & .05 &  .04 & .04 & .05 \\
& $\neq$ SotA & .32 & .36 & .25 & .44 & .45 & .18 & .46 & .48 & .16 & .32 & .25 & .33 &  .31 & .17 & .35 \\
& $\neq$ Top3 & .54 & .42 & .30 & .48 & .54 & .30 & .51 & .54 & .33 & .54 & .41 & .46 &  .39 & .26 & .39 \\
\midrule
\multirow{3}{*}{TC} 
& Disagree. &  .26 & .22 & .34 & .24 & .19 & .24 & .27 & .28 & .22 & .28 & .19 & .29 &  .18 & .24 & .20 \\
& $\neq$ SotA & .53 & .43 & .66 & .52 & .46 & .40 & .53 & .63 & .45 & .63 & .33 & .53 &  .30 & .40 & .27 \\
& $\neq$ Top3 & .57 & .60 & .63 & .57 & .56 & .60 & .62 & .55 & .47 & .63 & .60 & .60 &  .53 & .57 & .57 \\
\midrule
\multirow{3}{*}{PC} 
& Disagree. &  .28 & .24 & .32 & .25 & .23 & .22 & .21 & .22 & .22 & .12 & .20 & .19 &  .13 & .12 & .13 \\
& $\neq$ SotA & .50 & .50 & .63 & .42 & .53 & .43 & .28 & .33 & .30 & .33 & .47 & .50 &  .30 & .37 & .43 \\
& $\neq$ Top3 & .33 & .33 & .43 & .42 & .60 & .55 & .37 & .72 & .63 & .23 & .30 & .27 &  .27 & .20 & .07 \\
\midrule
\multirow{3}{*}{MSCOCO} 
& Disagree. &  .20 & .18 & .12 & .18 & .14 & .03 & - & - & - & - & - & - & - & - & - \\
& $\neq$ SotA & 1.0 & 1.0 & .00 & .03 & .03 & .00 & - & - & - & - & - & - & - & - & - \\
& $\neq$ Top3 & 1.0 & 1.0 & .17 & 1.0 & 1.0 & .47& - & - & - & - & - & - & - & - & - \\
\bottomrule
\end{tabular}
}
\caption{Disagreement between aggregation mechanisms per dataset and per metric.}
\label{tab:breakdown_disagreement}
\end{table*}

\section{Different view on uncertainty}
\label{app:uncertainty}
As argued in the main paper ( \Secref{ssec:stat_test}), the choice of aggregation mechanism bears strong similarities with the choice of statistical test. Thus, we measure in how many setups difference between systems that are statistically significant according to one test are also significant according to another.

% We study the pairs of systems on which the aggregation mechanisms agree, and ask whether different mechanisms provide the same view on uncertainty. Specifically, we wonder how often difference that are statistically significant for one mechanism are also statistically significant for another.

We compare: paired t-test (usually to compare means), the Mood's median test, and the sign test (consistent with BT). We also
% As measure of uncertainty, we use statistical testing as described above.
% For comparing means, we use: paired t-test, for comparing median, we use: Mood's median test, and for comparing results consistent with BT, we use: the sign test. We also
add the Wilcoxon sign-rank test as it was often recommended by previous work \cite{Owczarzak:2012,dror-etal-2018-hitchhikers}.

\begin{figure}
    \centering
    \includegraphics[width=0.55\linewidth]{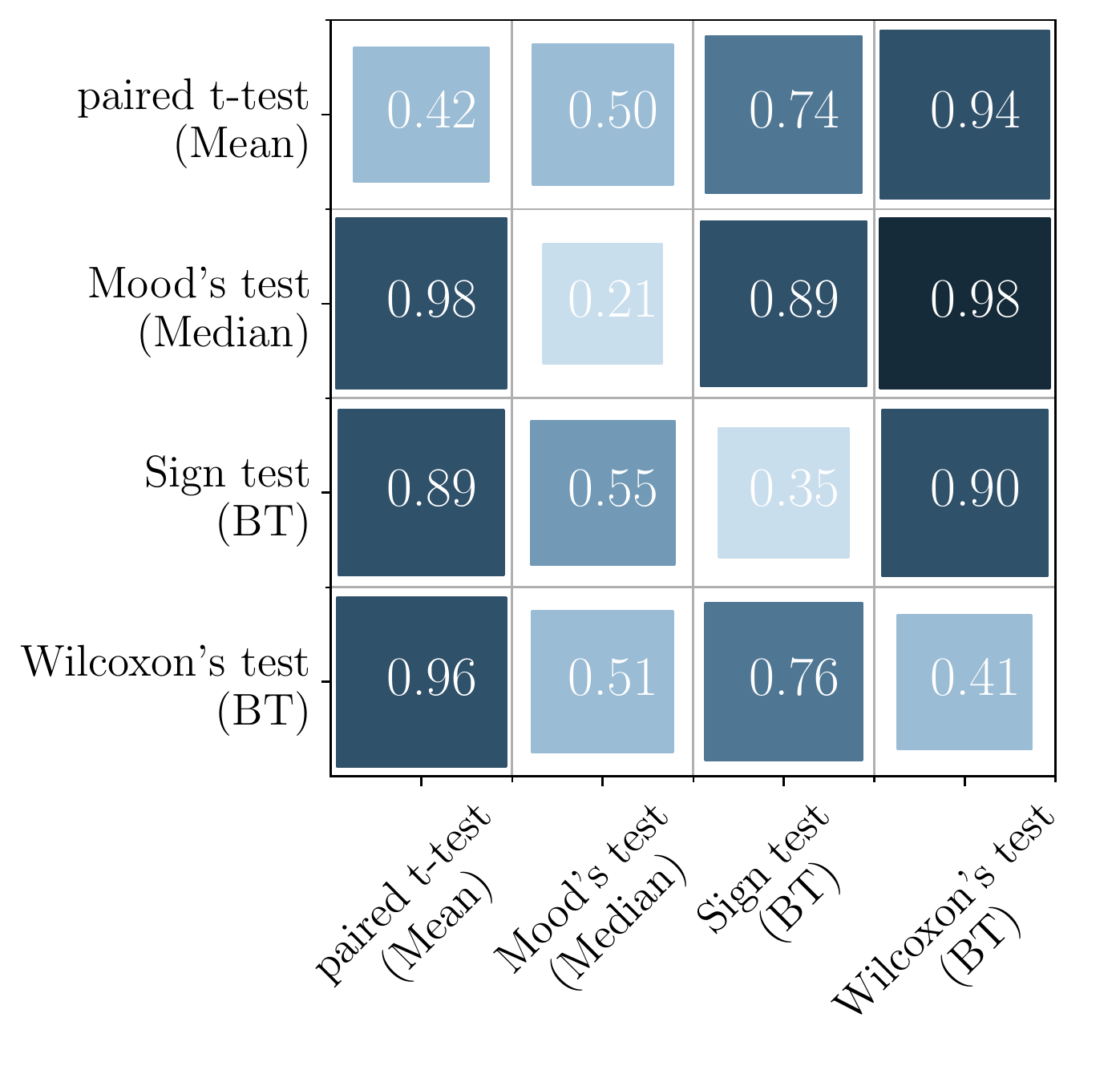}
    \caption{
    In this matrix, the cell in row $i$ and column $j$ indicates the frequency with which the test $j$ finds a difference significant among the pairs of systems for which the test $i$ has found the difference significant. For example, when the Mood's median test finds a significant difference between a pair, $98\%$ of the times, the paired t-test also finds the difference significant.
    % On the diagonal, we report the overall percentage of pairs on which the test finds the  differences to be significant.
    }
    \label{fig:uncertainty}
\end{figure}

In \Figref{fig:uncertainty}, we plot the frequency with which test $j$ yields  a  significant difference  among  the  pairs  of  systems  for which  the test $i$ has already yielded a significant difference.
The diagonal depicts the overall percentage of pairs of systems for which the test finds a significant difference. Note that the matrix is not symmetric.

Interestingly, %we observe that 
when the Mood's median test says the difference between two system is significant, $98\%$ of the times it is also the case for the paired t-test and $89\%$ of the times it is also the case for the Sign test.
So the Mood's median is the most restrictive, finding less often significant difference than the other two. In comparison, the Sign test and the Wilcoxon's sign-rank test find significant differences between systems much more frequently. In general, the paired t-test is the one finding differences the most frequently.
% In general, median and BT find significant differences less often than what is the current standard: the mean.

\section{Details about the Bradley--Terry model}
\label{app:bt_details}
Given a pair of systems $S_i$ and $S_j$, the Bradley--Terry model estimates the probability $p_{i,j}$ that the system $S_i$ is better than the system $S_j$ based on their relative strengths: $\frac{\lambda_i}{\lambda_i + \lambda_j}$.
% \begin{equation}
%     p_{i,j} = \mathbb{P}(S_i \geq S_j) = \frac{\lambda_i}{\lambda_i + \lambda_j},
% \end{equation}
% where $S_i > S_j$ means that the system $S_i$ receives a higher score than the system $S_j$ on a test instance drawn from some task (as evaluated by some fixed evaluation metric).

% Then, we also observe that:
% \begin{equation}
%     p_{j,i} = \mathbb{P}(S_j \geq S_i) = 1 - \mathbb{P}(S_i \geq S_j) = 1 - p_{i,j}
% \end{equation}

\cpt{BT} estimates these parameters $\lambda_i$ for each of the $n$ systems from the observed results of evaluation.
% , where for each test instance there are $n \choose 2 $ $= \frac{n(n-1)}{2}$ between the $n \choose 2$ pairs of systems.
We denote as $\omega_{i,j}$ the number of instances for which $S_i$ scores higher than $S_j$. Note that, in our setup, there is one comparison per test instance.
% Necessarily, in our case, we have $\omega_{i,j} + \omega_{j,i}=n$.
In the main paper, we said that the solutions for $\hat{\lambda}$ are found in closed-form for $n=2$.
% When the number of systems $n=2$, the parameters $\lambda_{i}$ can be found in closed-form:
% \begin{align}
%     \lambda_i = \omega_{i,j} \\
%     \lambda_j = \omega_{j,i}
% \end{align}
When the number of systems is greater than $2$, the parameters are found by an iterative optimization algorithm that maximizes the following log-likelihood:
\begin{equation}
    \mathcal{L}(\bm{\lambda}) = \sum\limits_{i=1}^n \sum\limits_{j=1}^n \omega_{i,j} \log(\lambda_i) - \omega_{i,j} \log(\lambda_i + \lambda_j),
\end{equation}
where $\boldsymbol{\lambda} = [\lambda_1, \dots, \lambda_n]$.

Denote $W_i$ as the number of comparison in which system $i$ is better: $W_i = \sum_j \omega_{i,j}$. Then, the algorithm iteratively performs the following two updates (at step $t$):
\begin{align}
& \hat{\lambda}_i = W_i \left( \sum\limits_{i \neq j} \frac{\omega_{i,j} + \omega_{j,i}}{\lambda_i^{(t)} + \lambda_j^{(t)}} \right)^{-1}, \ \forall i, \\
& \lambda_i^{(t+1)} = \frac{\hat{\lambda}_i}{\sum_k \hat{\lambda}_k}, \ \forall i.
\end{align}
It can be shown that starting from a random $\lambda$ this algorithm improves the log-likelihood at every iteration and converges to a unique maximum.

For the practical implementation, only a threshold $\epsilon$ defining when to stop has to be decided. We choose to stop iterating when at step $t$, if the new vector of parameter $\lambda$ remains close to the previous one: $\|\lambda^{(t+1)} - \lambda^{(t)}\|^2 < \epsilon$. Throughout our experiments, we always set $\epsilon = 1\cdot 10^{-9}$.

\section{Transitivity with BT and Arrow's theorem}
\label{app:arrow_theorem}
One possibly counter-intuitive behaviour of BT is that adding or removing a baseline can impact the scores and ordering of other systems. For example, consider two systems $A$ and $B$ with the following scores: $\mathcal{M}_A = [1, 2, 3]$ and $\mathcal{M}_B = [2, 3, 1]$. Then, \cpt{BT} identifies system $B$ as better with a relative strengths of $\frac{2}{3}$.
Now suppose another system $C$ is added with scores $\mathcal{M}_C = [3, 2, 1]$, running \cpt{BT} on these $3$ systems together gives the result that all systems have an equal strength, so now $B$ is not seen as better than $A$.

% To alleviate this issue, we propose to look at both the global results (\cpt{BT} on all systems available) and at all pairwise comparison: \cpt{BT}(A,B), \cpt{BT}(B, C), and \cpt{BT}(A,C). The result of \cpt{BT} for pairwise comparison of systems is easy to interpret and intuitive.

We search for triple of systems which exhibit this pattern in our data and couldn't find any as long as we use more than 10 test instance.

% However, the example presented above has a very pathological pairing structure, is it likely to occur in real data? We search for triples of systems such that like above one is deemed better in the pairwise comparison and then becomes equal or worse when the third system is added. We didn't such cases.

\paragraph{Can we hope to fix this weakness?}
Arrow's impossibility theorem says no \cite{arrow1950difficulty}. Our setup matches very well the problem of aggregating social preferences from voters. In this context, \newcite{arrow1950difficulty} proved that no aggregation mechanism with more than 2 voters and 3 possibilities can simulataneously meet the 3 following criterion: (i) monotonicity: if every voter prefers $X$ over $Y$, then the aggregation ranks $X$ above $Y$, (ii) (IIA) the aggregated preference between $X$ and $Y$ should remain unchanged if voter preferences between other pairs change, and (iii) no dictators: the outcome is not decided by a single voter.
% If there are more than 2 voters and more than $n=3$ possibilities, there exists no aggregation mechanism that can simultaneously meet the following three criterion:
% \begin{itemize}
% \denselist
%     \item (Monotonicity) If every voter prefers $X$ over $Y$, then the aggregation ranks $X$ above $Y$.
%     \item (Independence of Irrelevant Alternatives) the aggregated preference between $X$ and $Y$ should remain unchanged if voter preferences between other pairs change (like if the preference between $X$ and $Z$ changes for some voters)
%     \item (No dictators) The outcome is not decided by a single voter
% \end{itemize}
In our framework, voters are test instance and preferences are given by the evaluation metrics.
\cpt{BT} can fail on the second criteria, and \cpt{mean} and \cpt{MEdian} can be dictatorial (as seen in the paper).
% We just showed above that \cpt{BT} is failing on the second criteria and meeting the other two.
% Note that both \cpt{Mean} and \cpt{Median} fail on the third criteria and meet the other two. We provided examples in the main paper. 
A way around this problem is to remain with pairwise comparisons of systems $n<3$ and use \cpt{BT}. In that case, there is no possibility for \cpt{BT} to fail on IIA. 
% We provide this possibility in the tool we release.
% (see \Appref{app:tool})

% \section{A Simpson's paradox perspective}
% \label{app:simpson}
% A particular problem with the mean is well-known as the Simpson's Paradox. For example, consider two treatments $A$ and $B$ whose effectiveness are evaluated on  

\section{Variants of BT: Elo and TrueSkill}
\label{app:variants_bt}
\cpt{BT} has been extended in various ways. We discuss here two important variants that we incorporate in our analysis tool: Elo and TrueSkill.

\subsection{Elo ratings}
The Elo rating \cite{elo1978rating} is variant of the BT with an online update rule, i.e., the rating of systems (players) is updated as new test instances (new games) arrive.
As \cpt{BT}, Elo computes the probability that systems $S_i$ beats system $S_j$.
% \begin{align}
%     &\mathbb{P}(S_i \geq S_j) = \frac{Q_i}{Q_i + Q_j} = \frac{1}{1 + 10 ^{\frac{(R_i - R_j)}{400} }}, \\
%     &Q_k = 10^{\frac{R_k}{400}}, \ k \in \{i, j\},
% \end{align}
% where $Q_k$ plays a role analogous to $\lambda_k$ in BT and the parametrization of $\mathbb{P}(S_i \geq S_j)$ uses the logistic function. Then, $R_k$ is the rating of system $S_k$.
Now, the $t$-th test instance arrives and system $S_i$ receives the score $s_i$ and system $S_j$ receives the score $s_j$. 
% Based on the current ratings of systems, we expect the difference in score to be $\delta_{i,j} = \frac{s_i}{s_i + s_j} \approx \frac{Q_i}{Q_i + Q_j}$, and 
We update the rating $R$ based on this observed difference $\delta_{i,j}$:
\begin{equation}
    R_k^{(t+1)} = R^{(t)} + K \left(\delta_{i,j} - \frac{Q_i}{Q_i + Q_j}\right),
\end{equation}
where $K$ is parameter that has to be chosen, $R$ the rating of some system, and $Q$ plays a role analogous to $\lambda_k$ in \cpt{BT}.
% In chess, it is often set between $10$ and $40$. 
% The more the observed scores deviate from the expectation based on current ratings, the more the ratings change. 
$K$ controls how much each new instance can change the ratings.
It can be shown that, implicitly, Elo corresponds to a version of \cpt{BT} where the strength of systems is represented by a normal distribution: $\lambda_i + \epsilon_i, \ \ \epsilon_i \sim \mathcal{N}(0, \sigma^2)$, with a variance $\sigma^2$ shared by all players \cite{elo1978rating}.
% Elo can suffer from biases when players selectively choose their opponents or whether they participate or not in a tournament. This is not an issue in our setup because we can match-up all systems for all instances.
In our implementation, we provide the user with the ability to choose $K$ and set it to $20$ by default.

\subsection{TrueSkill}
TrueSkill \cite{true_skill} is Bayesian variant of the Elo rating system. It also updates the ratings of systems online, i.e., ratings change as new test instances arrive.
Now, the strength of a system $S_i$ is represented by a normal distribution, $\mathcal{N}(\lambda_i, \sigma_i^2)$. In contrast to Elo, each player has its own variance. The update follows Bayes rule, but is intractable in general, so message passing approximation are often employed.

% \newcite{true_skill} showed better predictive abilities compared to Elo for ranking video game players.

\section{Comparison of Elo, TrueSkill, and BT}
\label{app:repeating_experiments}
% Since Elo and TrueSkill are variants of BT, one might question which one is more suited to our problem setting. Elo and TrueSkill can model more complex scenario but at the cost of introducing new parameters, while BT remains simple and without parameters.

% The main difficulty with Elo and TrueSkill is to appropriately select the parameters. Elo has 1 parameter $K$, and TrueSkill has $5$ parameters.

% \subsection{Parameter variance}

% \subsection{Repeating experiments with Elo and TrueSkill}
We repeat the experiments of \Tabref{tab:global_results} from the main paper by replacing \cpt{BT} with Elo and TrueSkill with their default parameters.
The results are shown in \Tabref{tab:global_results_repeat}.
With Elo and TrueSkill, the same conclusions from the main paper hold, i.e., paired aggregation mechanisms exhibit significant disagreement with \cpt{MEAN} and \cpt{MEDIAN}.
Some discrepancies between \cpt{BT}, Elo, and TrueSkill remain which calls for further investigations about which one to choose.

\begin{table}
\centering
\resizebox{0.70\columnwidth}{!}{
\begin{tabular}{ l | c c c }
\toprule
& Disagree. & $\neq$ SotA & $\neq$ Top-3 \\
\midrule
\cpt{Mean} vs. \cpt{Median}   & $4\%$ & $18\%$ & $30\%$ \\
\cpt{Mean} vs. \cpt{BT}       & $9\%$ & $40\%$ & $49\%$ \\
\cpt{Median} vs. \cpt{BT}     & $9\%$ & $41\%$ & $55\%$ \\
\cpt{Mean} vs. Elo       & $20\%$ & $55\%$ & $84\%$ \\
\cpt{Median} vs. Elo     & $19\%$ & $56\%$ & $84\%$ \\
%% correlation 0.58
\cpt{Mean} vs. TrueSkill       & $18\%$ & $44\%$ & $76\%$ \\
\cpt{Median} vs. TrueSkill     & $17\%$ & $46\%$ & $79\%$ \\
\cpt{BT} vs. Elo        & $16\%$ & $38\%$ & $75\%$ \\
\cpt{BT} vs. TrueSkill  & $18\%$ & $53\%$ & $72\%$ \\
Elo vs. TrueSkill   & $18\%$ & $45\%$ & $71\%$ \\
\bottomrule
\end{tabular}
}
\caption{Global disagreement (as in 
\Tabref{tab:global_results}) between aggregation mechanisms repeated with Elo and TrueSkill.}
\label{tab:global_results_repeat}
\end{table}

\end{document}